\documentclass[a4paper,fleqn]{cas-dc}
\usepackage[numbers]{natbib}
\usepackage{times}
\usepackage{epsfig}
\usepackage{graphicx}
\usepackage{amsmath}
\usepackage{xpatch}
\usepackage{amssymb}
\usepackage{booktabs}
\usepackage{caption}
\usepackage{multirow}
\usepackage{mathtools}
\usepackage[utf8]{inputenc}

\usepackage{float}
\usepackage{multicol}
\usepackage{makecell}
\usepackage{rotating}
\usepackage{subcaption}
\usepackage{eucal}
\usepackage{mathtools}
\usepackage[english]{babel}
\usepackage{placeins}
\usepackage{stfloats}

\def\tsc#1{\csdef{#1}{\textsc{\lowercase{#1}}\xspace}}
\tsc{WGM}
\tsc{QE}
\tsc{EP}
\tsc{PMS}
\tsc{BEC}
\tsc{DE}

\begin{document}

\sloppy
\let\WriteBookmarks\relax
\def\floatpagepagefraction{1}
\def\textpagefraction{.001}
\shorttitle{A Comprehensive Study of Class Incremental Learning Algorithms for Visual Tasks}
\shortauthors{E. Belouadah et~al.}


\title [mode = title]{A Comprehensive Study of Class Incremental Learning Algorithms for Visual Tasks}                      




\author[1,2]{Eden Belouadah}[orcid=0000-0002-3418-1546]
\cormark[1]
\ead{eden.belouadah@cea.fr}


\address[1]{Université Paris-Saclay, CEA, Département Intelligence Ambiante et Systèmes Interactifs, 91191 Gif-sur-Yvette, France}

\author[1]{Adrian Popescu}[orcid=0000-0002-8099-824X]
\ead{adrian.popescu@cea.fr}

\author[2]{Ioannis Kanellos}[orcid=0000-0001-5323-1601]
\ead{ioannis.kanellos@imt-atlantique.fr}


\address[2]{IMT Atlantique, Computer Science Department, CS 83818 F-29238, Cedex 3, Brest, France}



\begin{abstract}
The ability of artificial agents to increment their capabilities when confronted with new data is an open challenge in artificial intelligence. 
The main challenge faced in such cases is catastrophic forgetting, i.e., the tendency of neural networks to underfit past data when new ones are ingested. 
A first group of approaches tackles forgetting by increasing deep model capacity to accommodate new knowledge.
A second type of approaches fix the deep model size and introduce a mechanism whose objective is to ensure a good compromise between stability and plasticity of the model. 
While the first type of algorithms were compared thoroughly, this is not the case for methods which exploit a fixed size model.
Here, we focus on the latter, place them in a common conceptual and experimental framework and propose the following contributions: 
(1) define six desirable properties of incremental learning algorithms and analyze them according to these properties, 
(2) introduce a unified formalization of the class-incremental learning problem,
(3) propose a common evaluation framework which is more thorough than existing ones in terms of number of datasets, size of datasets, size of bounded memory and number of incremental states, 
(4) investigate the usefulness of herding for past exemplars selection, 
(5) provide experimental evidence that it is possible to obtain competitive performance without the use of knowledge distillation to tackle catastrophic forgetting
and 
(6) facilitate reproducibility by integrating all tested methods in a common open-source repository.
The main experimental finding is that none of the existing algorithms achieves the best results in all evaluated settings. 
Important differences arise notably if a bounded memory of past classes is allowed or not.

\end{abstract}




\maketitle

\section{Introduction}
\label{intro}

Artificial agents which evolve in dynamic environments should be able to update their capabilities in order to integrate new data.
Depending on the work hypotheses made, names such as continual learning~\cite{seff2017continual,DBLP:conf/nips/ShinLKK17}, lifelong learning~\cite{DBLP:conf/cvpr/AljundiCT17,DBLP:journals/corr/abs-1802-07569} or incremental learning (IL)~\cite{deesil,DBLP:conf/eccv/CastroMGSA18,DBLP:conf/bmvc/He0SC18,DBLP:conf/cvpr/RebuffiKSL17} are used to describe associated works. 
The challenge faced in all cases is catastrophic forgetting~\cite{mccloskey:catastrophic}, i.e., the tendency of a neural network to underfit past data when new ones are ingested.
The effect of catastrophic forgetting is alleviated either by increasing the model capacity to accommodate new knowledge or by storing exemplars of past classes in a bounded memory and replaying them in each new state. 
Continual and lifelong learning algorithms usually increase model capacity and are tested in a setting in which a new task is added in each new state of the system. 
Recent comparative studies~\cite{lange2019, DBLP:journals/corr/abs-1802-07569} provide good coverage of these two types of approaches but give little room to incremental learning algorithms. 
We focus on class IL algorithms which provide the most interesting results in recent works from literature~\cite{belouadah2019il2m, scail2020,DBLP:conf/eccv/CastroMGSA18,DBLP:journals/corr/abs-1910-02509,DBLP:journals/corr/abs-1909-01520,DBLP:conf/cvpr/HouPLWL19,DBLP:conf/cvpr/RebuffiKSL17,DBLP:conf/cvpr/WuCWYLGF19}. 
Their study is interesting because one early example of such work~\cite{DBLP:conf/cvpr/RebuffiKSL17} is shown to outperform continual learning approaches when tested in a common experimental setting~\cite{lange2019}.
More recent works~\cite{belouadah2019il2m, scail2020,DBLP:conf/eccv/CastroMGSA18,DBLP:conf/cvpr/HouPLWL19,DBLP:conf/cvpr/WuCWYLGF19} have provided strong improvements compared to~\cite{DBLP:conf/cvpr/RebuffiKSL17}. 
We propose a series of contributions to better understand and evaluate existing class IL algorithms as well as interesting combinations of their components.

%

We first define a common analysis framework made of six desirable properties of incremental learning algorithms. 
This set of properties builds on the one proposed in~\cite{DBLP:conf/cvpr/RebuffiKSL17}, which already includes three of them (marked with * below):  
\begin{enumerate}
    \item \textbf{Complexity (C)*} - capacity to integrate new information with a minimal change in terms of the model structure. For a deep neural network, only the size of the classification layer should grow. Otherwise, the total number of parameters of the model is likely to increase strongly, especially at large scale.  
    \item \textbf{Memory (M)*} - ability to work with or without a fixed-size memory of past classes. Naturally, algorithms that do not require past data are preferable, but their performance is likely to be lower, especially if complexity growth is minimized.
    \item \textbf{Accuracy (A)*} - performance for past and new classes should approach that of a non-incremental learning process that has access to all data at all times.
    \item \textbf{Timeliness (T)} - delay needed between the occurrence of new data and their integration in the incremental models. 
    \item \textbf{Plasticity (P)} - capacity to deal with new classes that are significantly different from the ones that were learned in the past~\cite{DBLP:journals/corr/abs-1712-09708}.
    \item \textbf{Scalability (S)} - the aptitude to learn a large number of classes, typically up to tens of thousands, and ensure usability in complex real-world applications.
\end{enumerate}

Second, we propose a unified formalization of the class incremental learning problem and use it to analyze algorithms and results. 
Focus is put on the components which differentiate algorithms one from another in order to facilitate the understanding of advantages and limitations brought by each one of them. 
Moreover, we introduce promising combinations of components from different algorithms and assess their merits experimentally.

Third, we propose a thorough evaluation framework. 
Four public datasets designed for different visual tasks are used to test performance variability. 
Three splits in terms of the number of incremental states and three sizes for past memory are tested to assess performance robustness for these IL parameters, which were previously identified as being the most important~\cite{DBLP:conf/eccv/CastroMGSA18,DBLP:conf/cvpr/RebuffiKSL17}. 
We also propose an evaluation of the case when no past memory is allowed because this setting has a strong influence on algorithm performance.

Fourth, we examine the role of herding-based exemplar selection for past classes. Introduced in~\cite{DBLP:conf/icml/Welling09} and first used in an IL context by~\cite{DBLP:conf/cvpr/RebuffiKSL17}, its usefulness was questioned in~\cite{DBLP:conf/eccv/CastroMGSA18,DBLP:journals/corr/abs-1807-02802,liu2020mnemonics} where it was reported to provide only marginal improvement compared to random selection.
We run extensive experiments with the two selection algorithms and conclude that herding is useful for all methods tested.

Fifth, we show that it is possible to obtain interesting performance without the widely used knowledge distillation component~\cite{DBLP:conf/eccv/CastroMGSA18,DBLP:conf/cvpr/HouPLWL19,DBLP:journals/corr/abs-1807-02802,liu2020mnemonics,DBLP:conf/cvpr/RebuffiKSL17,DBLP:conf/cvpr/WuCWYLGF19}.
Instead, we use vanilla fine-tuning as a backbone for class IL with memory and model the problem as a case of imbalanced learning. 
The well-known thresholding method~\cite{DBLP:journals/nn/BudaMM18} is used to reduce the classification bias between past and new classes. 


Last but not least, we integrate the tested methods into a common open-source repository. 
We notably modify the implementations to make inputs and outputs uniform.
These changes will facilitate future experiments in the existing setting and will allow an easy extension to other datasets.

The main experimental finding here is that none of the existing class IL algorithms is better than the others in all experimental configurations. 
We find that both memory and incremental state sizes influence the relative performance of algorithms.
However, the most significant performance changes appear when testing IL with and without a memory of past classes.
These findings indicate that class incremental learning remains an open research problem, and further research efforts should be dedicated to it.

\begin{table*}\centering
\begin{center}
\resizebox{0.99\textwidth}{!}{

    \begin{tabular}{|p{0.04\textwidth}|p{0.32\textwidth}|p{0.32\textwidth}|p{0.32\textwidth}|}
    \toprule
    &  Model-Growth based & Fixed-Representation based & Fine-Tuning based\\
    \midrule
    \rotatebox[origin=b]{90}{\hspace{-5em} Complexity} 
    & The model evolves by adding parameters and weights to interconnect them~\cite{DBLP:conf/cvpr/RebuffiBV18,DBLP:journals/corr/RusuRDSKKPH16,DBLP:conf/cvpr/WangRH17} or small networks~\cite{DBLP:conf/cvpr/AljundiCT17} to include new knowledge. 
    The challenge is to optimize the effect of model growth on performance~\cite{lange2019,DBLP:conf/cvpr/RebuffiBV18}.
    & The model is fixed after the first non-incremental step. In a basic setting~\cite{deesil,DBLP:conf/nips/RebuffiBV17}, the only parameters added are those needed for new classes weights. In a more advanced setting~\cite{DBLP:conf/iclr/KemkerK18}, additional parameters are needed to improve past classes performance.
    & This group of IL methods are designed to work with a fixed structure of the backbone model. The number of parameters is only marginally affected by the modifications of the classification layer designed to reduce imbalance between past and new classes~\cite{DBLP:conf/eccv/CastroMGSA18,DBLP:conf/cvpr/HouPLWL19,DBLP:conf/cvpr/WuCWYLGF19}.
    \\
    \hline
   \rotatebox[origin=b]{90}{\hspace{-6em} Memory} 
   & The model growth allows for the deployment of these methods without the use of an exemplar memory. Memory is allocated to additional model parameters and weights instead of raw data for past classes, which is a more parsimonious way to store information about past classes~\cite{DBLP:conf/eccv/AljundiBERT18,DBLP:conf/cvpr/AljundiCT17, DBLP:journals/nn/RoyPR20}. 
   & Fixed-representations do not update the model during the incremental learning process and thus have a very low dependency on the memory of past classes~\cite{deesil}. Class weights are learned when they are first encountered and can be used throughout all subsequent incremental states. 
   & Performance of these methods is heavily dependent on the size of the past memory. However, storing a large amount of past exemplars is contradictory to IL objectives. Memory needs are reduced by exploiting knowledge distillation~\cite{DBLP:conf/eccv/CastroMGSA18,DBLP:conf/cvpr/HouPLWL19,DBLP:journals/corr/abs-1807-02802,DBLP:conf/cvpr/WuCWYLGF19} or by exploiting statistical properties of past states~\cite{belouadah2019il2m,scail2020}.
   \\
   \hline
    \rotatebox[origin=b]{90}{\hspace{-8em} Accuracy} 
   
   & Performance is correlated to the amount of model growth allowed. If growth is limited, MG-based methods have lower performance compared to that of FT-based ones~\cite{lange2019}. If significant growth is allowed significantly~\cite{DBLP:journals/nn/RoyPR20}, performance comes close to that of classical learning, but this is somewhat contradictory to the requirement to keep models complexity close to constant.
    & Accuracy is lower compared to FT-based methods because the model is not updated incrementally. High performance can be obtained with fixed-representations if the initial model is learned with a large dataset~\cite{deesil}, but the existence of such a dataset is a strong assumption in an incremental setting. 
     & Recent approaches report strong performance gain compared to previous work such as~\cite{DBLP:conf/eccv/CastroMGSA18,DBLP:journals/corr/abs-1807-02802,DBLP:conf/cvpr/RebuffiKSL17} either through more sophisticated definitions of knowledge distillation~\cite{DBLP:conf/cvpr/HouPLWL19} or through the casting of IL as an imbalanced learning problem~\cite{belouadah2019il2m,scail2020} or a combination of both~\cite{DBLP:conf/cvpr/WuCWYLGF19}. The gap with classical learning is narrowed if enough memory of past classes is allowed~\cite{DBLP:conf/cvpr/HouPLWL19,DBLP:conf/cvpr/WuCWYLGF19}.
    \\
    \hline
   \rotatebox[origin=b]{90}{\hspace{-5em} Timeliness} 
   & The complexity of model growth is generally similar to that of FT-based methods since retraining is needed for each incremental update~\cite{lange2019}.
   & Only the classifier weights layer needs to be trained and new knowledge is integrated in a timely manner~\cite{DBLP:journals/corr/abs-1909-01520}.   
   & New classes are not recognizable until retraining is finished to include them in the model. If applications are time-sensitive, an acceleration of the training process can be envisioned at the expense of result suboptimality.
   \\
   \hline
   \rotatebox[origin=b]{90}{\hspace{-4em}Plasticity} 
   & MG-based methods are specifically designed to cope with different visual tasks~\cite{DBLP:conf/eccv/AljundiBERT18,DBLP:conf/eccv/MallyaDL18}. The challenge is to minimize the amount of additional parameters needed to accommodate each new task~\cite{lange2019}.
   & Plasticity is limited since the representation is learned in the first state and then fixed. Performance drops significantly if the incremental tasks change a lot and the initial representation is not transferable anymore~\cite{DBLP:conf/cvpr/RazavianASC14}.
   & The model updates enable adaptation to new data as they are streamed into the system. If no memory is allowed, plasticity is too important and this shift is controlled through knowledge distillation or imbalance handling.
   \\
   \hline
   \rotatebox[origin=b]{90}{\hspace{-8em}Scalability} 
   & These methods scale well to new classes or tasks as long as the systems in which they are deployed have sufficient resources to support the underlying model growth for training and inference phases, as well as for its storage.
   & The dependence on the bounded memory is limited and FR-based methods can include a very large number of classes. This is possible because class weights are learned in their initial state and reused later. 
   & The size of the bounded memory determines the number of past classes for which exemplars can be stored and which are still recognizable when new ones are integrated.
   If the memory constraints allow for this, the memory can be increased to keep the number of exemplars per class constant~\cite{DBLP:conf/cvpr/HouPLWL19}.
   \\
   \hline
   \hline
   \rotatebox[origin=b]{90}{\hspace{-18em}Global assessment} 
   & Approaches in this group cope well with new data, are not or weakly dependent on a memory of the past, and are scalable to a large number of classes. 
However, complexity is a disadvantage since the model has to grow in order to integrate new knowledge.
They also require retraining when new classes are added, and timeliness is not optimal.  
They are usable when: model complexity can grow during the incremental process, streamed data vary a lot between incremental states, no storage is available for past data and immediate use of updated models is not essential. 
   & Fixed-representation methods inherit the advantages and disadvantages of transfer learning schemes.
Model complexity is constant and they can be updated in a timely manner since only the classification layer is retrained.
They have a low dependency on past memory and can scale up to a large number of classes. 
However, these algorithms are heavily dependant on the quality of the initial representation and have a low plasticity. 
They are usable when: model complexity should stay constant, data variability is low, no storage is available for past memory and updates are needed in a timely manner. 
   & Fine-tuning based methods are adequate when we try to optimize the architecture complexity and the plasticity~\cite{DBLP:journals/corr/abs-1712-09708} of representations. 
However, since they require network retraining when new classes are added, their timeliness is not optimal. 
Equally important, the bounded memory constraint makes it hard to scale because the memory will eventually become too small to represent past classes adequately. 
They are usable when: model complexity should stay constant, streamed data vary a lot between incremental states, storage is available for past data and immediate use of updated models is not essential. 
   \\        
    \bottomrule
    \end{tabular}
}
\end{center}
    \caption{Analysis of the main groups of incremental learning algorithms with respect to their desirable properties. A global assessment with recommended use cases is also provided.}
    \label{tab:il_analysis}
\end{table*}

\section{Related work}
\label{sec:sota}
There is a strong regain of interest for incremental learning due to the proposal of different deep learning based algorithms.
We categorize recent approaches in three main groups and map each group to the six IL properties from Section~\ref{intro} in Table~\ref{tab:il_analysis}. 
We discuss the advantages and/or challenges related to each group-property pair to facilitate their comparison.
We also provide a global assessment with a focus on the application contexts in which each type of approach could be deployed.


\textbf{Model-Growth (MG)} based methods increase the size of deep models to include new knowledge. 
Wang et al.~\cite{DBLP:conf/cvpr/WangRH17} introduced \textit{Growing a Brain}, a method based on increasing representational capacity by widening or deepening the network.
\textit{Progressive Neural Networks}~\cite{DBLP:journals/corr/RusuRDSKKPH16} are an alternative approach that exploits several models during training to preserve knowledge from past tasks. 
Lateral connections between all models are learned to leverage prior knowledge of past features and thus reduce the effect of catastrophic forgetting.
Recently, ~\cite{DBLP:journals/nn/RoyPR20} propose an adaptive network that enables self-growth in a tree-like manner. It is based on features hierarchy reorganization whenever new tasks arrive.

Aljundi et al.~\cite{DBLP:conf/cvpr/AljundiCT17} present a lifelong learning architecture based on a network of experts.
A gating mechanism uses training samples to decide which expert to transfer knowledge from for each task. 
\textit{Deep Adaptation Networks}~\cite{DBLP:journals/corr/RosenfeldT17} is another model-growth based approach which adds additional parameters for each new task. The architecture is importantly augmented if a large number of new tasks arrives.
The approach presented in~\cite{DBLP:conf/cvpr/RebuffiBV18} is based on several neural networks that share the majority of parameters and add modular adapters to connect the networks and specialize each one of them for a specific task.

$PackNet$~\cite{DBLP:conf/cvpr/MallyaL18} is based on a pruning technique that identifies redundant parameters and uses them to train the network on new tasks. The approach is not able to learn a large number of tasks since the network can not be strongly compressed without significant performance loss.
$Piggyback$~\cite{DBLP:conf/eccv/MallyaDL18} is a modified version of $PackNet$ that exploits network quantization to propose masks for individual weights.
It thus learns a large number of tasks with a single base network. The approach increases the model complexity as extra parameters are added to include new tasks. 
Alternately, \textit{Memory Aware Synapses (MAS)}~\cite{DBLP:conf/eccv/AljundiBERT18} deploys a mechanism that identifies the most important weights in the model by looking at the sensitivity of the output function instead of the loss. When a new task arrives, changes to important weights are penalized. This method is basically designed to work with unlabeled datasets, but was later adapted for usage with unlabeled datasets~\cite{DBLP:conf/cvpr/AljundiKT19}. 
This use case is very interesting but, for now, limited to specific tasks such as face recognition.

\textit{Self-Organizing Maps} (SOMs) are online unsupervised learning algorithms that rely on approximate stochastic gradient technique, and can be adapted to Incremental Learning. 
\textit{Neural Gas} (NG) networks~\cite{ng_1993} and its growing NG variant~\cite{fritzke1994growing} are related to SOMs which are often exploited for incremental learning.
\textit{PROjection-PREdiction} ($PROPRE$)~\cite{som_2017} is an incremental learner based on NG and SOMs, which implements an extra supervised read-out layer implemented as a linear regression, as well as a concept drift detection mechanism in order to make the SOM usuable in IL context.  
\textit{Neural Gas with local Principal Component Analysis} ($NGPCA$)~\cite{gas_il_2010} is an online incremental learner focused on robot platform for object manipulation tasks. 
It modifies the classical NG algorithm to extend nodes to ellipoids to better match the data distribution. 
\textit{Dynamic Online Growing Neural Gas} ($DYNG$)~\cite{DYNG_2013} is an online classification approach that controls the growing speed of the NG network in such a way to speed up learning for new knowledge while slowing down the growth for the already learned knowledge. 
$NGPCA$ and $DYNG$ are very interesting methods but not directly comparable to the methods evaluated here since they do not exploit deep learning backbones.
NG and SOM are growing networks that were widely used for incremental semi-supervised clustering~\cite{gas_som_1994}, multi-class online classification problems~\cite{gas2}, and online semi-supervised vector quantization learning~\cite{DBLP:journals/ijon/ShenZZL20}. 

We note that SOM and NG are originally designed for unsupervised learning and an adaptation to a supervised scenario is needed for comparability with the approaches which are in focus here.
\textit{TOpology-Preserving knowledge InCrementer} ($TOPIC$)~\cite{fscil_2020} is a very recent such work which adapts NG to class incremental learning for visual datasets with focus on few-shot learning, with a method named $FSCIL$, but experiments are also run for the standard scenario.
First, $TOPIC$ introduces an NG network to learn feature space topologies for knowledge representation.
The network grows to learn new classes while also dealing with changes in the feature space due to deep model update. 
This is achieved using a \textit{Min-Max} loss that pushes new classes that share the same label to a new NG node, while pulling the new nodes of different labels away from each others. 
Second, $TOPIC$ preserves past knowledge by stabilizing the topology of the NG network using an \textit{Anchor Loss} term. 
Since $TOPIC$ focuses on the feature space which encodes more semantic information than the raw classification scores, it is less affected by the bias induced by high new classes' raw scores.
A topology-preserving network named $TPCIL$ is introduced in~\cite{DBLP:conf/eccv/TaoCHWG20} to handle catastrophic forgetting. 
The network models the feature space using an Elastic Hebbian Graph, and the topology is maintained using a topology-preserving loss that constrains the neighborhood relationships in the graph when learning new classes. 
This approach augments the Hebbian graph by inserting vertices for each new class.
The addition of nodes in $TOPIC$ and $TPCIL$ gradually increases the complexity of the architecture.

\textit{Incremental Learning Vector Quantization} ($ILVQ$)~\cite{DBLP:conf/pakdd/XuFHZ09} is a prototype-based classifier that does not require prior knowledge of the number of prototypes or their initial value. Instead, it uses a threshold-based insertion scheme, based on training data distribution to determine the number of required prototypes for each class. The main drawback of this approach is the continuously increasing architecture to store the learned patterns in order to function in an IL setting.

\textbf{Fixed-Representation (FR)} based methods do not update the deep representation for each incremental state and are less present in literature. 
They can be seen as a basic variant of fine-tuning based methods.
A fixed-representation method is briefly described in~\cite{DBLP:conf/cvpr/RebuffiKSL17}. 
The results reported with it are poor, and this is due to a suboptimal usage of the method.
In particular, the classification layer for past classes is needlessly relearned in each incremental state using only the exemplars of each class.
Since they rely on a fixed representation, the stronger classifier weights learned initially with all past class data are reusable.
\textit{Deep Shallow Incremental Learning} ($DeeSIL$)~\cite{deesil} is a method which applies a simple transfer learning scheme~\cite{DBLP:journals/corr/abs-1805-08974, DBLP:conf/cvpr/RazavianASC14}. 
The approach makes use of a deep fixed representation to learn the first batch of classes and a battery of \textit{Support Vector Machines} (SVMs)~\cite{DBLP:conf/colt/BoserGV92} to incrementally learn new classes. 

\textit{FearNet}~\cite{DBLP:conf/iclr/KemkerK18} is a biologically inspired such method. 
Separate networks are used for long and short term memories to represent past and new classes. 
A decision mechanism is implemented to decide which network should be used for each test example.
$FearNet$ is interesting, but its memory increases significantly with time since the algorithm needs to store detailed statistics for each class learned.

\textit{Deep Streaming Linear Discriminant Analysis} ($Deep$-$SLDA$)~\cite{DBLP:journals/corr/abs-1909-01520} is an online approach based on SLDA~\cite{DBLP:journals/tsmc/PangOK05} algorithm. The Network is trained on the first batch of classes and is frozen afterwards. During training, a class-specific running mean vector and a shared covariance matrix are updated, while the prediction is done by assigning the label to the closest Gaussian in feature space defined by the class-mean vectors and covariance matrix.

\textit{REplay using Memory INDexing} ($REMIND$)~\cite{DBLP:journals/corr/abs-1910-02509} is brain inspired by the hippocampal indexing theory.
The method is also based on an initial representation which is only partially updated afterwards.
The approach uses a vector quantization technique to stores compressed intermediate representations of images, which are more compact than images. The stored vectors are reconstructed and replayed for memory consolidation. Vector quantization is widely used in unsupervised incremental learning~\cite{DBLP:journals/pr/Lughofer08}. Here, the authors combine the \textit{Adaptive Resonance Theory} (ART) with variant of vector quantization to balance the trade-off between plasticity and stability during incremental online learning. This approach was designed to handle two- and high-dimensional data within image classification framework. We tackle in this paper supervised learning, and this approach is not compatible with our experimental protocol.

\textbf{Fine-Tuning (FT)} based methods form a group which often uses a distillation~\cite{DBLP:journals/corr/HintonVD15} term to reduce catastrophic forgetting~\cite{mccloskey:catastrophic}.
The use of knowledge distillation in an IL context is similar to self-distillation~\cite{DBLP:conf/icml/FurlanelloLTIA18,DBLP:conf/cvpr/YimJBK17} in that it operates with the same network architecture for the teacher and the student.
However, a notable difference arises from the fact that new data are progressively incorporated.
\textit{Learning without Forgetting} ($LwF$)~\cite{DBLP:conf/eccv/LiH16} is a pioneering work that does not require a memory of past classes. 
It leverages knowledge distillation~\cite{DBLP:journals/corr/HintonVD15} to minimize the discrepancy between representations of past classes from the previous and current IL states. 
$LwF$ first performs a warm-up step that freezes the past parameters and trains only the new ones and then jointly trains all network parameters until convergence.

\textit{Incremental Classifier and Representation Learning} ($iCaRL$)~\cite{DBLP:conf/cvpr/RebuffiKSL17} is a popular IL method that combines the use of $LwF$ and of a memory for past class exemplars storage.
Classification is performed with a nearest-mean-of-exemplars method instead of the raw scores predicted by the network. 
This external classifier is deployed to reduce the prediction bias in favor of new classes, which occurs due to data imbalance between past and new classes.
An iCaRL analysis~\cite{DBLP:journals/corr/abs-1807-02802} concludes that its most important components are the fixed-size memory and the distillation loss. 
The herding mechanism and the nearest-mean-of-exemplars classification seem to matter less. 
The authors of ~\cite{DBLP:conf/eccv/CastroMGSA18} present an IL algorithm which differs from $iCaRL$ mainly through the way prediction bias is reduced. 
The external classifier is replaced by a balanced fine-tuning step, which uses the same number of samples for past and new classes.
This component has an important impact on performance and leads to a strong improvement compared to $iCaRL$.
Sophisticated data augmentation is also used and has a small positive influence on results. 

\textit{Learning without Memorizing} ($LwM$)~\cite{DBLP:journals/corr/abs-1811-08051} is a distillation based approach that does not need memory for past classes. Instead, the authors propose an information preserving penalty using attention distillation loss that captures the changes in the classifier attention maps in order to preserve past knowledge. 
In~\cite{DBLP:conf/wacv/ZhangZGLTHZK20}, another distillation based system is proposed, it trains two separate networks, one for new classes and one for past classes, and then combines them via a double distillation loss. A deep memory consolidation is also performed using unlabeled auxiliary data to replace past class memory.

A part of recent IL approaches focuses on a more sophisticated tackling of catastrophic forgetting. 
The authors of \textit{Multi-model and Multi-level Knowledge Distillation} ($M2KD$)~\cite{DBLP:journals/corr/abs-1904-01769} propose a loss that distills knowledge not only from the previous model but from all the past models where the classes have been learned for the first time. 
They also propose an additional distillation term that operates on the intermediate layers of the CNN in addition to the last fully connected one.  
In~\cite{DBLP:journals/access/XiangMCX20}, knowledge distillation is also combined with a fixed-size memory of the past. The authors deploy an algorithm to set a dynamic vector which corrects the bias induced by distillation loss among past classes and improves the representativeness of past image features.
Hou et al.~\cite{DBLP:conf/cvpr/HouPLWL19} present \textit{Learning a Unified Classifier Incrementally via Rebalancing} ($LUCIR$), a method which gains a lot of traction.
$LUCIR$ is based on three main components: (1) \textit{cosine normalization} balances the magnitudes of past and new class probabilities, (2) \textit{less-forget constraint} modifies the usual distillation loss to handle feature vectors instead of raw scores and (3) \textit{inter-class separation} encourages the network to separate past and new class embeddings and actually implements a theoretical finding from~\cite{DBLP:conf/icml/PhuongL19}.
Note that further improvement of IL with distillation could be obtained by adapting recent theoretical and empirical advances such as those described in ~\cite{DBLP:conf/cvpr/ParkKLC19} and ~\cite{DBLP:conf/icml/PhuongL19}. Alternately, $PODNet$~\cite{DBLP:conf/eccv/DouillardCORV20} relies on a spacial-based distillation loss that constrains the evolution of the model's representation, and multiple proxy vectors to flexibly represent learned classes. This approach is more adequate with long runs of small incremental tasks.

Another recent stream of research focuses on modeling IL as an imbalanced learning problem. 
\textit{Bias Correction} ($BiC$)~\cite{DBLP:conf/cvpr/WuCWYLGF19} is a recent approach that uses a classical knowledge distillation term and adds a linear layer after the prediction layer of the deep model to reduce the bias in favor of new classes. 
The layer needs a validation set to learn parameters and is effective as long as the size of the validation set is sufficient. 
\textit{Class Incremental Learning with Dual Memory} ($IL2M$)~\cite{belouadah2019il2m} advocates for the use of vanilla fine-tuning as a backbone for IL. 
A very compact memory that stores classification statistics from the initial state of each classifier is added.
Its content is leveraged to rectify scores of past classes and make them more comparable to those of new classes. 
\textit{Maintaining Discrimination and Fairness} ($MDF$)~\cite{mdf_2020} is very similar to $IL2M$. The main difference is that $MDF$ keeps the distillation loss to maintain discrimination between past classes. In $MDF$, the rectification of class scores is done by aligning new class weights to those of past classes by multiplying each new class weight by the mean norm of past class weights and dividing it by the mean norm of new class weights, before to finally compute the prediction scores.

\textit{Classifier Weights Scailing for Class Incremental Learning} ($ScaIL$)~\cite{scail2020} is motivated by the same hypothesis as $IL2M$~\cite{belouadah2019il2m}, $MDF$~\cite{mdf_2020} and $BiC$~\cite{DBLP:conf/cvpr/WuCWYLGF19}. 
Inspired by fixed-representation methods, bias reduction is achieved by reusing the classifier weights learned initially with all data. 
The experimental results indicate that, while the model evolves throughout incremental states, initial classifiers are still usable after a scaling operation, which makes them comparable to classifiers learned for new data. 
\textit{Standardization of Initial Weights} (SIW)~\cite{siw_20200} is also based on initial weights replay in a memoryless IL setting. The weights replay is followed by standardization of all class weights to smooth weights distribution in order to tackle catastrophic forgetting.

\textit{Mnemonics Training}~\cite{mnemonics_2020} is built on top of herding-based approaches such as $iCaRL$, $BiC$ and $LUCIR$ to modify the herding procedure by parameterizing exemplars and making them optimizable. The network is then optimized in two manners: model-level and exemplar-level. The memory is thus adjusted incrementally to match the data distribution in an effective way, leading mnemonic exemplars to yield separation between classes.
In embedding systems, \textit{Semantic Drift Compensation} ($SDC$)~\cite{sdc_2020} was proposed to estimate the semantic drift of past knowledge while learning new knowledge to compensate for it, to further improve performance. The drift is computed at the class-mean-embedding level, which means that this approach is based on NCM classifier that does not need exemplars storage, since the past class-mean embeddings are estimated using new data only.
In~\cite{singh2020calibrating}, authors propose an approach that calibrates activation maps of the CNN in order to accommodate new knowledge. Calibration is done using spatial and channel-wise calibration modules, and only the calibration parameters are trained at each new incremental state. This method does not require a past-class memory. However, the calibration parameters grow instead.

Using \textit{Generative Adversarial Networks} (GANs) to generate past data holds promise since it reduces the memory footprint of algorithms.
However, despite recent progress~\cite{Ghosh_2018_CVPR}, generated images are still sub-optimal for IL. 
Since additional GAN models need to be created, the complexity in the number of parameters is fair but not optimal.
The authors of~\cite{DBLP:journals/corr/abs-1802-00853} use a GAN to create artificial images for past classes. Generated and real examples are mixed to obtain slightly better performance than that of $iCaRL$~\cite{DBLP:conf/cvpr/RebuffiKSL17}. However, the performance significantly drops when relying exclusively on artificially generated images. Alternately, \textit{GAN Memory with No Forgetting}~\cite{DBLP:journals/corr/abs-2006-07543} is based on sequential style modulations to represent the past memory by forming a sequential targeted generative models. Here, the memory itself is designed as a form of lifelong learning.  

This paper is focused on a scenario that requires a constant complexity of deep models and investigate the effect of allowing a fixed-size memory or not. Consequently, experiments are conducted with approaches that are fit to work under these conditions, namely those based on fine tuning and fixed representations.

\section{Problem formalization}
We propose a formalization of class incremental learning which builds on those introduced in~\cite{belouadah2019il2m,scail2020,DBLP:conf/eccv/CastroMGSA18,DBLP:conf/nips/RebuffiBV17}.
Given an initial non-incremental state $\mathcal{S}_0$, a model $\mathcal{M}_0$ is trained from scratch on a dataset $\mathcal{D}_0 = \{(X_0^j, Y^j_0); j=1,2, ..., P_0 \}$. $X_0^j$ and $Y_0^j$ are respectively the set of images and labels for the $j^{th}$ class in $\mathcal{S}_0$, $N_0 = P_0$ is the number of classes in the first non-incremental state. 

We note $T$ the total number of states, including the initial state and $T-1$ incremental states.  
A new batch of $P_t$ new classes is streamed in each incremental state $\mathcal{S}_t$ and the objective is to learn a model $\mathcal{M}_{t}$ which recognizes $N_t=P_0 + P_1 + ... + P_t$ classes.
This model is trained using the previous state model $\mathcal{M}_{t-1}$ on a dataset $\mathcal{D}_t = \{(X_t^j, Y^t_0); j=1,2, ..., P_t \} \cup \mathcal{K}$.
Note that all data of the new $P_t$ classes are available with only a bounded exemplar subset $\mathcal{K}$ of data from the $N_{t-1}=P_0 + P_1 + ... + P_{t-1}$ past classes.
An imbalance in favor of new classes appears and grows throughout incremental states since the bounded memory $\mathcal{K}$ needs to be allocated to a larger number of past classes each time.

As discussed in Section~\ref{sec:sota}, recent class incremental learning algorithms were implemented using deep convolutional networks (DNNs) as a backbone. 
While DNNs are end-to-end classification approaches, a part of the IL algorithms use a separate classifier layer. 
In such cases, the model $\mathcal{M}_t$ includes two main components: a feature extractor $\mathcal{F}_t$ and a classification component $\mathcal{C}_t$.

The feature extractor $\mathcal{F}_t$ is defined as:

\begin{equation}
\begin{aligned}
\mathcal{F}_t : X_t \rightarrow \mathbb{R}^d \hspace{3em}\\
\textbf{x} \mapsto \mathcal{F}_t(\textbf{x})=\boldsymbol{f}^{\textbf{x}}_t
\end{aligned}
\end{equation}

where $\boldsymbol{f}^{\textbf{x}}_t$ is a $d$-dimensional compact vectorial representation of the image $\textbf{x}$.

The classifier $\mathcal{C}_t$ is usually defined as:

\begin{equation}\label{eq:raw_scores}
\begin{aligned}
    \mathcal{C}_t :  \mathbb{R}^d &\rightarrow&  \mathbb{R}^{N_t} \hspace{10em} \\
   \boldsymbol{f}^{\textbf{x}}_t &\mapsto& \mathcal{C}_t(\boldsymbol{f}^{\textbf{x}}_t)= \boldsymbol{f}^{\textbf{x}}_t \times \boldsymbol{W}_t + \boldsymbol{b}_t = \boldsymbol{o}_t
\end{aligned}
\end{equation}

where: 
\begin{itemize}
 \item $\boldsymbol{o}_t=(o_t^1, o_t^2,..., o_t^{N_t})$ is the vector of raw scores of size $N_t$ providing the individual prediction scores for each class $j=1, 2, ..., N_t$. 
 
 \item $\boldsymbol{W}_t$  and $\boldsymbol{b}_t$ are the weights matrix and bias vector of the last fully connected layer of size $(d, N_t)$ and $N_t$ respectively.  The size $d$ of the feature vector depends on the CNN architecture used.
\end{itemize}

In an end-to-end DNN~\cite{DBLP:conf/cvpr/HeZRS16}, a softmax function is applied to transform raw scores $\boldsymbol{o}_t=(o_t^1, o_t^2,..., o_t^{N_t})$ to probabilities $\boldsymbol{p}_t=(p_t^1, p_t^2,..., p_t^{N_t})$.
This is the case for approaches such as ~\cite{belouadah2019il2m,DBLP:conf/eccv/CastroMGSA18,DBLP:conf/cvpr/HouPLWL19}. 
$\mathcal{C}_t$ can also be implemented using an external classifier. 
For instance,~\cite{DBLP:conf/cvpr/RebuffiKSL17} and a variant of~\cite{DBLP:conf/cvpr/HouPLWL19} use a Nearest-Class-Mean (NCM) as a classifier to compute class prediction based on the similarity of each test image to the average class feature computed from the available images in each state.
Yet another choice~\cite{deesil} is to exploit a fixed deep representation to extract features for all incremental states and a battery of linear SVMs to implement $\mathcal{C}$.  

\subsection{Loss function}
A majority of existing IL algorithms~\cite{DBLP:conf/eccv/CastroMGSA18,DBLP:conf/cvpr/HouPLWL19,DBLP:journals/corr/abs-1807-02802,DBLP:conf/cvpr/WuCWYLGF19} alleviate the effects of catastrophic forgetting~\cite{mccloskey:catastrophic} by introducing a distillation term in the loss function~\cite{DBLP:conf/eccv/CastroMGSA18,DBLP:conf/cvpr/RebuffiKSL17}.
In general, this function takes the following form:

\begin{equation}
\mathcal{L} = \lambda~\times \mathcal{L}^c +(1-\lambda)~\times \mathcal{L}^d
\end{equation}

where $\mathcal{L}^c$ and $\mathcal{L}^d$ are classical cross-entropy and distillation terms, respectively. $\lambda \in [0, 1]$ is a hyper-parameter that provides the weight of each loss term.

\textit{Cross-Entropy Loss} in the state $\mathcal{S}_t$ is computed for all past and new classes and is given by :

\begin{equation}
    \mathcal{L}_t^c(\textbf{x}) = \sum_{(\textbf{x},y)~\in~\mathcal{D}_t \cup \mathcal{K}}~\sum_{j=1}^{N_t} - \mathbb{1}_{y=j}~log[p_t^j(\textbf{x})]
\end{equation}

where $\mathbb{1}$ is the indicator function.

\textit{Distillation Loss} in the state $\mathcal{S}_t$ is computed for past classes and is given by:

\begin{equation}
    \mathcal{L}_t^d(\textbf{x}) = \sum_{(\textbf{x},y)~\in~\mathcal{D}_t\cup \mathcal{K}}~\sum_{j=1}^{N_{t-1}} - \Phi_{t-1}^j(\textbf{x}) ~log[\Phi_t^j(\textbf{x})]
\end{equation}

where $\Phi$ is the softened softmax applied on the raw scores predicted by the network. 
The softened score of the $j^{th}$ class in state $\mathcal{S}_t$ is:

\begin{equation}
    \Phi_t^j(\textbf{x}) = \frac{e^{o^j_t(\textbf{x}) / \mathcal{T}}}{\sum_{l=1}^{N_t} e^{o_t^l(\textbf{x}) / \mathcal{T}} }
\end{equation}

where $\mathcal{T}$ is a temperature scalar. 
$\mathcal{L}^d$ was originally introduced to improve performance when no memory of past classes is available~\cite{DBLP:conf/eccv/LiH16}.
The authors of~\cite{DBLP:conf/nips/RebuffiBV17} adapted it for the case when a bounded memory $\mathcal{K}$ is allowed. 
Different flavors of distillation were later proposed in~\cite{DBLP:conf/eccv/CastroMGSA18,DBLP:conf/cvpr/HouPLWL19,DBLP:journals/corr/abs-1807-02802,DBLP:conf/cvpr/WuCWYLGF19}.

\subsection{Score bias correction}
Recent works~\cite{belouadah2019il2m,scail2020, DBLP:conf/cvpr/WuCWYLGF19} hypothesize that due to the limited memory of the past, IL is akin to imbalanced learning where the network is trained with enough images for the new classes but only a few ones for past classes.

These works handle catastrophic forgetting as an imbalance issue and thus make use of an additional bias removal layer $\mathcal{R}_t$ at the end of the model $\mathcal{M}_t$. The layer takes the raw scores $\boldsymbol{o}_t$ predicted by the classifier $\mathcal{C}_t$ and multiply those of new classes by a reducing factor to make them more comparable to those of past classes, giving a chance to the latter to be selected during inference. 

\begin{equation}\label{eq:bias_removal_layer1}
\begin{aligned}
    \mathcal{R}_t :  \mathbb{R}^{N_t} &\rightarrow&  \mathbb{R}^{N_t} \hspace{11.5em} \\
   \boldsymbol{o}_t &\mapsto& \mathcal{R}_t(\boldsymbol{o}_t)= [r(o_t^1),.., r(o_t^j),..,r(o_t^{N_t})]
\end{aligned}
\end{equation}

where:

\begin{equation}\label{eq:bias_removal_layer2}
    r(o_t^j) =
\left\{
    \begin{array}{ll}
        o_t^j  & \mbox{if }~~~~~~~~~ 1 \leq j \leq N_{t-1} \\
        \alpha o_t^j + \beta & \mbox{if }~~ N_{t-1} < j \leq N_t
    \end{array}
\right.
\end{equation}

\vspace{1em}

$\alpha$ and $\beta$ are scaling factors. They differ from an approach to another. $BiC$~\cite{DBLP:conf/cvpr/WuCWYLGF19} learns them using a validation set after fine-tuning the model. $IL2M$~\cite{belouadah2019il2m} uses past classes' statistics to set the value of $\alpha$ while zeroing $\beta$, and the rectification aims to increase past classes' scores instead of decreasing those of new classes. $MDF$~\cite{mdf_2020} aligns new classes' weights with those of past classes by setting $\alpha$ to  the mean norm of past class weights divided by the mean norm of new class weights. Finally,
$ScaIL$~\cite{scail2020} normalizes the weights matrix of $\mathcal{C}_t$ leading to an implicit change in the values of the scores $\boldsymbol{o}_t$.

\subsection{Past memory management}
The bounded memory $\mathcal{K}$, which stores a partial view of past classes, is a central component of existing IL algorithms. 
Regardless of the state $\mathcal{S}_t$, the same number of images $|\mathcal{K}_t^j|=|\mathcal{K}|/N_t$ is kept for each class   . 
Because the memory capacity is bounded, the number of images for each past class is reduced at the end of each state to accommodate images from new classes. 
The representation of past classes is thus degraded since the number of images per class is progressively reduced. 
Consequently, the model is biased toward new classes and underfits past classes, a well-known effect of catastrophic forgetting~\cite{mccloskey:catastrophic}. 

The role of exemplar selection techniques is debated in literature~\cite{DBLP:conf/eccv/CastroMGSA18,liu2020mnemonics,DBLP:conf/nips/RebuffiBV17,DBLP:conf/cvpr/WuCWYLGF19}.
Notably, the authors of~\cite{DBLP:conf/eccv/CastroMGSA18,liu2020mnemonics} report similar results with herding-based and random selection of exemplars.
We run extensive experiments with both methods and reach a different conclusion. 
This is a result of a different interpretation of herding definition.
Both~\cite{DBLP:conf/eccv/CastroMGSA18} and~\cite{liu2020mnemonics} select exemplars statically using their similarity to the mean embedding of the class.
We follow the original definition from~\cite{DBLP:conf/icml/Welling09}, also exploited in~\cite{DBLP:conf/cvpr/RebuffiKSL17}, and select each exemplar based on a dynamic mean computed at each step based on the exemplars which were already selected. 
The advantage of the original definition is that it provides a better approximation of the actual class center compared to a static selection. 
It consists in taking, for each class, the set of images having the closest mean to the real class mean. Exemplar means are computed on feature vectors extracted from the penultimate layer of the CNN. 
We also tried exemplar selection techniques inspired from active learning approaches such as: entropy~\cite{DBLP:journals/sigmobile/Shannon01}, min margin~\cite{Settles10activelearning}, core-set~\cite{DBLP:conf/iclr/SenerS18}, k-means~\cite{bodo2011}. Initial experiments showed that none of these techniques provide better performance than herding.
Consequently, we do not report such results.

\section{Fine-Tuning based IL algorithms}

\label{sec:methods_ft}

\begin{table*}
    \begin{center}
    \resizebox{\textwidth}{!}
    {
    \begin{tabular}{|c|c|c|c|c|c|c|c|c|c|c|c|c|c|c|c|c|}
        \hline
         \textbf{Characteristics} & \makecell{$LwF$\\ \cite{DBLP:conf/eccv/LiH16}} & \makecell{$LwF^{init}_{*}$\\ \cite{siw_20200}}& \makecell{$iCaRL$\\ \cite{DBLP:conf/cvpr/RebuffiKSL17}} & \makecell{$LUCIR$\\ \cite{DBLP:conf/cvpr/HouPLWL19}} &
         \makecell{$FT$\\~\cite{belouadah2019il2m,scail2020,DBLP:conf/cvpr/RebuffiKSL17}} & \makecell{$FT^{NEM}$\\~\cite{belouadah2019il2m,scail2020}} & \makecell{$FT^{BAL}$\\~\cite{belouadah2019il2m,scail2020}} & \makecell{$BiC$\\ ~\cite{DBLP:conf/cvpr/WuCWYLGF19}} & 
         \makecell{$ScaIL$\\\cite{scail2020}} & \makecell{$IL2M$\\ \cite{belouadah2019il2m}} & \makecell{$FT^{th}$\\\cite{DBLP:journals/nn/BudaMM18}} & \makecell{$FT^{init}_{*}$\\~\cite{siw_20200}} & 
         \makecell{$FR$\\~\cite{DBLP:conf/cvpr/RebuffiKSL17}} &
         \makecell{$DeeSIL$\\~\cite{deesil}} &
         \makecell{$REMIND$\\~\cite{DBLP:journals/corr/abs-1910-02509}} &
         \makecell{$Deep$-$SLDA$\\~\cite{DBLP:journals/corr/abs-1909-01520}}
         \\ \hline
         Model update & \checkmark  & \checkmark&  \checkmark  &  \checkmark  &  \checkmark  & \checkmark  & \checkmark  & \checkmark &  \checkmark  & \checkmark  & \checkmark  & \checkmark  & $\times$  & $\times$ & $\times$ & $\times$  \\ 
         Distillation &  \checkmark &  \checkmark& \checkmark  &  \checkmark  &  $\times$  & $\times$  & $\times$   & \checkmark &  $\times$  & $\times$  & $\times$  & $\times$  & $\times$  & $\times$ & $\times$ &  $\times$ \\ 
         Bias removal & $\times$ & \checkmark& \checkmark  &  \checkmark &  $\times$  & \checkmark  & \checkmark   & \checkmark & \checkmark  & \checkmark  & \checkmark  & \checkmark  & $\times$  & $\times$ & $\times$ & $\times$  \\ 
         Memory usage  & NC & NC&  C  &  NC  & NC  & C  & C & C  & C  & C  & C  & NC  & NC  & NC & NC & NC \\ \hline 

    \end{tabular}
    }
    \end{center}
    \caption{Main characteristics of tested approaches: model update - indicates if the model is trained for each incremental state; distillation - if this part of the loss is exploited to control catastrophic forgetting; bias removal - is a separate component which is specifically dedicated to balancing results between past and new classes, and memory usage - if this component is compulsory (C) or not (NC). * refers to all methods built on top of $init$.}
    \label{tab:tested_algos}
\end{table*}

We compare recent class incremental algorithms and also adaptations of them, which combine components from different algorithms. 
An overview of the tested algorithms and of their characteristics is presented in Table~\ref{tab:tested_algos}.
The model update for each incremental state is widely used in existing approaches.
Distillation is also used by a majority of algorithms from literature to counter the effect of catastrophic forgetting.
Bias removal aims at balancing predictions for past and new classes. 
It is deployed either as a complement to distillation~\cite{DBLP:conf/eccv/CastroMGSA18,DBLP:conf/cvpr/RebuffiKSL17,DBLP:conf/cvpr/WuCWYLGF19} or to replace it~\cite{belouadah2019il2m,scail2020}. 
Memory usage is compulsory for methods that rely heavily on exemplars of past classes.
The dominant approach is based on model updating via fine-tuning to integrate new knowledge~\cite{belouadah2019il2m,scail2020,DBLP:conf/eccv/CastroMGSA18,DBLP:conf/cvpr/HouPLWL19,DBLP:conf/cvpr/WuCWYLGF19}.
The performance of these algorithms depends heavily on the existence of a bounded memory of the past.
We also present results with fixed-representation based algorithms~\cite{deesil, DBLP:journals/corr/abs-1910-02509, DBLP:journals/corr/abs-1909-01520,DBLP:conf/cvpr/RebuffiKSL17}, which do not update the model but are also less dependent on memory.

\textbf{\textit{LwF}}~\cite{DBLP:conf/eccv/LiH16} uses distillation loss to encourage the model $\mathcal{M}_t$ to predict the same scores for past classes in the current state $\mathcal{S}_t$ than in the previous one. 
$LwF$ was designed to work without a memory of past classes.
\textit{LwF} constitutes the inspiration for all class IL algorithms~\cite{DBLP:conf/eccv/CastroMGSA18, DBLP:conf/cvpr/HouPLWL19,DBLP:journals/corr/abs-1807-02802,DBLP:conf/cvpr/WuCWYLGF19}.

\vspace{0.2cm}
\textbf{\textit{iCaRL}}~\cite{DBLP:conf/cvpr/RebuffiKSL17} exploits fine-tuning with distillation loss $\mathcal{L}^d$ to prevent catastrophic forgetting and a variant of Nearest-Class-Mean~\cite{DBLP:journals/pami/MensinkVPC13} to counter imbalance between past and new classes. The main difference with $LwF$ is the introduction of a bounded memory to enable efficient replay.
 
\vspace{0.2cm}        
\textbf{\textit{LUCIR}}~\cite{DBLP:conf/cvpr/HouPLWL19} is based on fine-tuning with an integrated objective function. 
Authors propose the following contributions: (1) cosine normalization to balance magnitudes of past and new classifiers (2) less forget constraint to preserve the geometry of past classes and (3) inter-class separation to maximize the distances between past and new classes. 
The combination of these contributions constitutes a more sophisticated take at countering catastrophic forgetting compared to the use of knowledge distillation from~\cite{DBLP:conf/eccv/CastroMGSA18,DBLP:journals/corr/abs-1807-02802,DBLP:conf/cvpr/RebuffiKSL17}. We experiment with two versions of this approach: 
\begin{itemize}
    \item $LUCIR^{NCM}$ - the original definition proposed by the authors where a Nearest-Class-Mean classifier is used. This version is functional in presence of a memory only, due to the need for exemplars to compute past class-mean..
    \item $LUCIR^{CNN}$ - the network outputs are used for classification. This version can be deployed with or without a memory for the past.
\end{itemize}

\vspace{0.2cm} 
\textbf{\textit{FT}}~\cite{belouadah2019il2m,scail2020,DBLP:conf/cvpr/RebuffiKSL17} is the plain use of vanilla fine-tuning. 
The model $\mathcal{M}_t$ is initialized with the weights of the previous model $\mathcal{M}_{t-1}$ and only the cross-entropy loss $\mathcal{L}^c$ is used. 
$FT$ constitutes the simplest way to update models in incremental learning. 
It is heavily affected by catastrophic forgetting if a bounded memory is not available~\cite{DBLP:conf/cvpr/RebuffiKSL17} but becomes an interesting baseline if a memory is available~\cite{belouadah2019il2m}.

\vspace{0.2cm}
\textbf{\textit{FT$^{NEM}$}}~\cite{belouadah2019il2m,scail2020} is a version of $FT$ which replaces the classifier $\mathcal{C}_t$ by a NEM classifier from ~\cite{DBLP:conf/cvpr/RebuffiKSL17}. $FT^{NEM}$ is a modified version of $iCaRL$ in which the distillation loss $\mathcal{L}^d$ is ablated. 
\textit{FT$^{NEM}$} can only be deployed if a bounded memory is allowed to alleviate catastrophic forgetting.
    
\vspace{0.2cm}
\textbf{\textit{FT$^{BAL}$}}~\cite{belouadah2019il2m,scail2020} is inspired by~\cite{DBLP:conf/eccv/CastroMGSA18}. 
A vanilla $FT$ is first performed, followed by a balanced $FT$.
This second step aims to reduce bias between past and new classes by training on a version of $\mathcal{D}_t$, which stores the same number of images for past and new classes to obtain similar magnitudes for them.
\textit{FT$^{BAL}$} is also tributary to the fixed memory $\mathcal{K}$ because past exemplars are needed for the balancing step.
In the absence of memory, this approach becomes equivalent to $FT$ and is heavily affected by catastrophic forgetting.
    
\vspace{0.2cm}
\textbf{\textit{BiC}}~\cite{DBLP:conf/cvpr/WuCWYLGF19} adds a linear layer for bias removal which is trained separately from the rest of the model learned with cross-entropy and distillation losses.
The objective of the supplementary layer is to reduce the magnitudes of predictions for new classes to make them more comparable to those of past classes.
Since a validation set is needed to optimize the bias removal layer, \textit{BiC} can only function if a memory $\mathcal{K}$ is available.
We note that $\mathcal{K}$ needs to be large enough to obtain reliable parameter estimations.


\vspace{0.2cm}
 \textbf{\textit{ScaIL}}~\cite{scail2020} hypothesizes that the classification layer $\mathcal{C}_t$ learned when classes were first streamed and learned with all data can be reused later.
 The main challenge is that deep models $\mathcal{M}$ are updated between incremental states.
 Normalization of the initial $\mathcal{C}_t$ is proposed to mitigate the effect of model updates and make past and new classes' predictions comparable.
 \textit{ScaIL} needs a bounded memory to keep trace of past classes in the embeddings.

\vspace{0.2cm}    
\textbf{\textit{IL2M}}~\cite{belouadah2019il2m} uses past classes' statistics to reduce the prediction bias in favor of new classes. 
Past classes' scores are modified using the ratio between their mean classification score when learned initially in the state $\mathcal{S}_i$ and in the current state $\mathcal{S}_t$.
Furthermore, the ratio between the mean classification score over all classes in $\mathcal{S}_t$ and $\mathcal{S}_i$ is also used.
This approach is also prone to catastrophic forgetting if no memory is allowed.

\vspace{0.2cm}    
\textbf{\textit{FT$^{th}$}}, inspired by imbalanced learning~\cite{DBLP:journals/nn/BudaMM18}, it implements fine tuning followed by threshold calibration (also known as threshold moving or post scaling). Thresholding adjusts the decision threshold of the model by adding a calibration layer at the end of the model during inference to compensate the prediction bias in favor of new classes:

   \begin{equation}
    {p^j_t}^\prime = p^j_t \times {\frac{|\mathcal{D}_t|}{|X_t^j|}} 
    \end{equation}
    
    where $|X_t^j|$ is the number of samples for the $j^{th}$ class and $|\mathcal{D}_t|$ is the total number of training examples in the state $\mathcal{S}_t$. A memory of the past is needed to rectify past classes' scores.
This method is also heavily dependant on the bounded memory.

\vspace{0.2cm}
\textbf{\textit{FT$^{init}$}}~\cite{scail2020}, \textbf{\textit{FT$^{init}_{L2}$}}~\cite{scail2020}, \textbf{\textit{FT$^{init}_{L2+mc}$}}~\cite{scail2020, siw_20200}, \textbf{\textit{FT$^{init}_{siw+mc}$}}~\cite{siw_20200}, \textbf{\textit{LwF$^{init}$}}~\cite{siw_20200},
\textbf{\textit{LwF$^{init}_{L2}$}}~\cite{siw_20200}, 
and \textbf{\textit{LwF$^{init}_{siw}$}}~\cite{siw_20200}, 
are methods built on top of $FT$ and $LwF$, in order to reduce the bias of the network towards new classes, where:

\begin{itemize}
    \item $init$ - replaces the weights of past classes in the current state with their initial weights learned in the initial state with all available data. 
    \item $L2$ - normalization that makes classifier weights more comparable across states.
    \item $siw$ - standardization of last layer weights, in order to make them comparable. 
    \item $mc$ - state mean calibration defined as:
    
       \begin{equation}
    {p_t^j}^\prime = p_t^j \times \frac{\mu(\mathcal{M}_t)}{\mu(\mathcal{M}^j_i)}
    \end{equation}
    
    $\mu(\mathcal{M}_t)$ and $\mu(\mathcal{M}^j_i)$ - means of top-1 predictions of models learned in the current state and the initial state of the $j^{th}$ class computed over their training sets.
    
\end{itemize}

The four components can be combined together, where first $init$ is applied, followed either by $L2$ or $siw$ normalization, and finally followed by the state mean calibration $mc$.
These methods are mostly interesting for IL without memory because they do not require the use of a past exemplars memory.

\section{Fixed-Representation based IL algorithms}
\label{sec:methods_fr}
\textbf{\textit{Fixed-Representation (FR)}}~\cite{DBLP:conf/cvpr/RebuffiKSL17} exploits the initial model $\mathcal{M}_0$ trained on the classes of $\mathcal{S}_0$ and freezes all its layers except the classification one in later incremental states. 
The frozen model is a limitation but also an advantage in that it allows the reuse of initial classifier layers, learned with all images, throughout the entire incremental process.
Unfortunately, the reuse of initial layers, is not done in ~\cite{DBLP:conf/cvpr/RebuffiKSL17} and results are suboptimal.
The method does not need a bounded memory for the update.
    
\textbf{\textit{DeeSIL}}~\cite{deesil} is a variant of $FR$ in which the classification layer of DNNs is replaced by linear SVMs.
\textit{DeeSIL} is a straightforward application of a transfer learning~\cite{DBLP:journals/corr/abs-1805-08974,DBLP:conf/cvpr/RazavianASC14} scheme in an incremental context. 
The use of external classifiers is proposed because they are faster to optimize compared to an end-to-end $FR$.

\textbf{\textit{Deep-SLDA}}~\cite{DBLP:journals/corr/abs-1909-01520} defines the model as $\mathcal{M}_t  \equiv  F(G(\cdot))$ where $G$ is the fixed upper part of the network and $F(\cdot)$ is the output layer. Only $F(\cdot)$ is trained across incremental states in a streaming manner,  while $G(\cdot)$ serves as a feature extractor. At each incremental state, $Deep$-$SLDA$ updates a class-specific running mean vector and a running shared covariance matrix among classes. During inference, it assigns an image to the closest Gaussian in feature space defined by the class mean vectors and the covariance matrix. $Deep$-$SLDA$ does not need to store past class data and it is thus functional in absence of memory.

\textbf{\textit{REMIND}}~\cite{DBLP:journals/corr/abs-1910-02509} shares the same definition of $\mathcal{M}_t$ as $Deep$-$SLDA$, where $G(\cdot)$ is the first fifteen convolutional and three down sampling layers, and $F(\cdot)$ is the remaining two convolutional and one fully connected layers of a ResNet-18~\cite{DBLP:conf/cvpr/HeZRS16}. $REMIND$ relies on Product Quantizer (PQ)~\cite{DBLP:journals/pami/JegouDS11} algorithm to store intermediate representations of images as compressed vectors for fast learning. The compact vectors are then reconstructed and replayed for memory consolidation. Note that compact vectors allow us to save much more past data than with raw images (for instance, all ILSVRC can fit in the memory when $|\mathcal{K}|=20000$).

\section{Experimental setup}
Experiments are done with all IL approaches presented in  Sections~\ref{sec:methods_ft} and~\ref{sec:methods_fr}.
We also provide results with $Full$, a classical non-incremental training from scratch where all classes are learned with all their data. 
This algorithm is the upper bound for all class incremental approaches.

\subsection{Datasets}
\label{sub:metho}
Four datasets designed for object, face, and landmark recognition are used here. 
The choice of significantly different tasks is important to study the adaptability and robustness of the tested methods.
The main dataset statistics are provided in Table~\ref{tab:dataset}.

\begin{itemize}
\item \textbf{ILSVRC} ~\cite{DBLP:journals/ijcv/RussakovskyDSKS15} is a subset of 1000 $ImageNet$ classes used in the $ImagenetLSVRC$ challenges. It is constituted of leaves of the $ImageNet$ hierarchy which most often depicts specific visual concepts.

\item \textbf{VGGFACE2}~\cite{DBLP:conf/fgr/CaoSXPZ18} is designed for face recognition. We selected 1000 classes having the largest number of associated images. Face cropping is done with MTCNN~\cite{DBLP:journals/spl/ZhangZLQ16} before further processing. 

\item \textbf{Google~Landmarks}~\cite{DBLP:conf/iccv/NohASWH17} (\textbf{LANDMARKS} below) is built for landmark recognition, and we selected 1000 classes having the largest number of associated images.

\item \textbf{CIFAR100}~\cite{Krizhevsky09learningmultiple} is designed for object recognition and includes 100 basic level classes~\cite{ROSCH1976382}. 

\end{itemize}

\begin{table}
\begin{center}
    \resizebox{0.47\textwidth}{!}
    {
    \begin{tabular}{|c|c|c|c|c|}
        \hline
         Dataset & train & test & $\mu$(train) & $\sigma$(train) \\ \hline
         ILSVRC     &  1,231,167  & 50,000  & 1231.16 & 70.18 \\ \hline 
         VGGFACE2   &   491,746  & 50,000  & 491.74 & 49.37 \\ \hline 
         LANDMARKS  &  374,367  & 20,000 & 374.36 & 103.82 \\ \hline 
         CIFAR100   &  50,000  & 10,000  & 500.00 & 0.00 \\ \hline 

    \end{tabular}
    }
    \end{center}
    \caption{Main statistics for the evaluation datasets, $\mu$ is the mean number of images per class; and $\sigma$ is the standard deviation of the distribution of the number of images per class.}
    \label{tab:dataset}
\end{table}

\subsection{Experimental protocol}
The experimental protocol is inspired by the one proposed in $iCaRL$~\cite{DBLP:conf/cvpr/RebuffiKSL17}. 
The most important parameters in IL are the number of states $T$, and the size of the memory $\mathcal{K}$, and we test three values for each one of them.
First, we fix the number of states $T=10$ and vary the memory to include approximately $2\%, 1\%, 0.5\%$ of the full training sets. 
Memory sizes are thus $|\mathcal{K}|=\{20000, 10000, 5000\}$ for ILSVRC, $|\mathcal{K}|=\{10000, 5000, 2500\}$ for VGGFACE2, $|\mathcal{K}|=\{8000, 4000, 2000\}$ for LANDMARKS and $|\mathcal{K}|=\{1000, 500, 250\}$ for CIFAR100.
Second, we fix the memory size $|\mathcal{K}|$ to include $0.5\%$ of the full training sets and vary the number of states $T=\{20, 50\}$. 
Here, we chose the smallest memory size because it represents the most challenging case when a memory is allowed. 
This is also the most interesting in practice since it requires a reduced resource to store past data.
We report results with $|\mathcal{K}|=0$ separately since the absence of memory renders some of the algorithms completely inoperable while others are still working.

\subsection{Implementation details}
A ResNet-18 architecture~\cite{DBLP:conf/cvpr/HeZRS16} with an SGD optimizer is used as a backbone for all the methods. 
$REMIND$~\cite{DBLP:journals/corr/abs-1910-02509}, $Deep$-$SLDA$~\cite{DBLP:journals/corr/abs-1909-01520}, 
$BiC$~\cite{DBLP:conf/cvpr/WuCWYLGF19} and $LUCIR$~\cite{DBLP:conf/cvpr/HouPLWL19} are run using the optimal parameters of the public implementations provided in the original papers. 
$iCaRL$~\cite{DBLP:conf/cvpr/RebuffiKSL17} is run using the code from~\cite{DBLP:conf/cvpr/HouPLWL19} since it provides better performance than the original implementation.
$LwF$~\cite{DBLP:conf/eccv/LiH16} is run using the code from~\cite{DBLP:conf/cvpr/RebuffiKSL17},
The SVMs in $DeeSIL$~\cite{deesil} are implemented using LinearSVC solver from Scikit-Learn toolbox~\cite{DBLP:journals/corr/abs-1201-0490}. The SVMs were optimized using classical grid search as described in the original paper. 

$FT$ and its derivatives are based on the same fine-tuning backbone and are implemented in Pytorch~\cite{paszke2017automatic}. 
Training images are processed using randomly resized $224\times224$ crops, horizontal flipping, and are normalized afterward.
Given the difference in scale and the number of images between CIFAR100 and the other datasets, we found that a different parametrization was needed for this dataset. 
Note that the parameters' values presented below are largely inspired by the original ones given in~\cite{DBLP:conf/cvpr/HeZRS16}.

For CIFAR100, the first non-incremental state and $Full$ are run for 300 epochs with $batch~size=128$, $momentum=0.9$ and $weight~decay=0.0005$. 
The $lr$ is set to 0.1 and is divided by 10 when the error plateaus for 60 consecutive epochs. 
The incremental states of $FT$ and $FR$ are trained for 70 epochs with $batch~size=128$, $momentum=0.9$ and $weight~decay=0.0005$. 
The learning rate is set to $lr=0.1 / t$ at the beginning of each incremental state $\mathcal{S}_t$ and is divided by 10 when the error plateaus for 15 consecutive epochs.
 
For ILSVRC, VGGFACE2 and LANDMARKS, the first non-incremental state and $Full$ are run for 120 epochs with $batch~size=256$, $momentum=0.9$ and $weight~decay=0.0001$. The $lr$ is set to 0.1 and is divided by 10 when the error plateaus for 10 consecutive epochs. 
The incremental states of $FT$ and $FR$ are trained for 35 epochs with $batch~size=256$, $momentum=0.9$ and $weight~decay=0.0001$. 
The learning rate is set to $lr=0.1 / t$ at the beginning of each incremental state $\mathcal{S}_t$ and is divided by 10 when the error plateaus for 5 consecutive epochs.

Dataset details and the code of all tested approaches and their adaptations are publicly available to facilitate reproducibility. \footnote{\url{https://github.com/EdenBelouadah/class-incremental-learning}}

\subsection{Evaluation measures}
The main evaluation measure used here is the popular top-5 accuracy~\cite{DBLP:journals/ijcv/RussakovskyDSKS15}. Following~\cite{DBLP:conf/eccv/CastroMGSA18}, accuracy is averaged only for incremental states (i.e., excluding the initial, non-incremental state), which is not of interest from an IL perspective. 
The sizes of the past memory $\mathcal{K}$ and the number of states $T$ are varied to evaluate the robustness of algorithms.  

Since a relatively large number of configurations is tested, it is convenient also to use a global measure.
We use the incremental learning gap measure ($G_{IL}$)~\cite{scail2020}, which computes the average performance gap between a classical learning and each IL configuration for each algorithm. 
$G_{IL}$ is defined as:

\begin{equation}
       G_{IL} = \frac{1}{Z} \times \sum_{z=1}^{Z} \frac{acc_z - acc_{Full}}{acc_{Max} - acc_{Full}} 
    \label{eq:gap}    
\end{equation}

where: $Z$ is the number of tested configurations; $acc_{z}$ is the top-5 score for each configuration; $acc_{Full}$ is the upper-bound accuracy of the dataset ($Full$ accuracy); $acc_{Max}$ is the maximum theoretical value obtainable for the measure ($acc_{Max} = 100$ here).

Following~\cite{DBLP:journals/corr/abs-1712-09708}, the denominator is introduced in order to ensure that no individual configuration has an exaggerated influence on the global score.
Note that $G_{IL}$ is related to the forgetting rate proposed in~\cite{liu2020mnemonics}, which is actually the numerator of individual configurations from Equation~\ref{eq:gap}.



\begin{table*}
\begin{center}
\resizebox{\textwidth}{!}{
\begin{tabular}{|c||c|c|c|c|c|c|c|c|c|c|c|c||cc|cc|cc|cc||c|}
\hline
States & \multicolumn{12}{c||}{$T=10$} & \multicolumn{8}{c||}{$|\mathcal{K}|=0.5\%$} & \multirow{3}{*}{$G_{IL}$}\\
\cline{1-21}
Dataset & \multicolumn{3}{c|}{ILSVRC} &\multicolumn{3}{c|}{VGGFACE2} & \multicolumn{3}{c|}{LANDMARKS} & \multicolumn{3}{c||}{CIFAR100} & \multicolumn{2}{c|}{ILSVRC} &\multicolumn{2}{c|}{VGGFACE2} & \multicolumn{2}{c|}{LANDMARKS} & \multicolumn{2}{c||}{CIFAR100} & \\
\cline{1-21}
$|\mathcal{K}|$ & $2\%$ & $1\%$ & $0.5\%$ & $2\%$ & $1\%$ & $0.5\%$ & $2\%$ & $1\%$ & $0.5\%$ & $2\%$ & $1\%$ & $0.5\%$ & T=20 & T=50 & T=20 & T=50 & T=20 & T=50 & T=20 & T=50 &   \\
\hline

{\small $iCaRL$}& 79.3 & 76.5 & 71.0 & 96.0 & 95.3 & 93.9 & 95.1 & 94.0 & 91.8& 66.5 & 56.1 & 47.9 & 55.9 & 45.0 & 88.5 & 78.2 & 86.8 & 82.4 & 35.5 & 35.4 & -7.36 \\
{\small $LUCIR^{CNN}$}& 79.9 & 76.4 & 72.6 & 97.2 & \textbf{96.9} & \textbf{96.5} & 97.2 & 96.6 & 96.1 & 79.8 & 75.4 & 69.9 & 63.9 & 55.3 & 93.5 & 88.3 & 93.7 & 90.5 & 53.5 & 47.9 & -4.13 \\
{\small $LUCIR^{NEM}$}& 80.5 & 80.0 & 79.4 & 96.2 & 96.0 & 95.7 & 95.4 & 94.9 & 94.4 & 82.6 & 80.8 & 78.8 & 73.6 & 66.3 & 92.7& 87.9 & 91.9 & 89.8 & 69.0 & 63.0 & -4.33 \\
{\small $FT$} & 79.4 & 74.4 & 65.9 & 96.4 & 94.5 & 91.3 & 96.6 & 94.7 & 91.4 & 82.4 & 77.9 & 70.7 & 69.4 & 64.3 & 91.6 & 89.2 & 90.9 & 89.0 & 64.3 & 54.8 & -5.19 \\
{\small $FT^{NEM}$ }& 81.4 & 79.0 & 75.0 & 96.4 & 95.4 & 94.0 & 96.1 & 94.6 & 92.6 & 85.1 & 81.7  & 76.0 & 76.5 & 69.0 & 94.0 & 91.1 & 91.9 & 89.9 & 68.8 & 55.9 & -4.28 \\
{\small $FT^{BAL}$}& 84.0 & 80.9 & 76.5& 97.0 & 95.7 & 92.4 & 96.9 & 95.3 & 92.2 & 80.0 & 74.0 & 69.0 & 75.9 & 67.1 & 92.3 & 89.5 & 91.2 & 88.9 & 62.9 & 54.2 & -4.70 \\
{\small $BiC$}& \textbf{85.5} & \textbf{82.8} & \textbf{79.7} & \textbf{97.3} & 96.6 & 95.7 & \textbf{97.9} & \textbf{97.3} & \textbf{96.6} & \textbf{88.8} & \textbf{87.6} & \textbf{83.5} & 74.6 & 63.9 & 92.3 & 85.3 & \textbf{94.7} & 90.5 & 50.5 & 19.6 & -4.03 \\
{\small $ScaIL$} & 82.0 & 79.8 & 76.6 & 96.5 & 95.8 & 95.2 & 97.3 & 96.0 & 94.0 & 85.6 & 83.2 & 79.1 & 76.6 & 70.9 & \textbf{95.0} & \textbf{92.4} & 92.6 & 90.4 & 69.8 & 51.0 & -3.70 \\
{\small $IL2M$ }& 80.9 & 78.1  & 73.9 & 96.7  & 95.4  & 93.4  & 96.5 & 94.7  & 92.5 & 81.8 & 77.0  & 71.2 & 70.9 & 60.6 & 92.5 & 88.4 & 90.8 & 88.1 & 61.5 & 51.0 & -4.95 \\
{\small $FT^{th}$}& 84.3 & 82.1 & 78.3 & 97.2 & 96.3 & 94.8 & 97.2 & 95.8 & 94.0 & 86.4 & 83.9 & 79.1 & \textbf{78.6} & \textbf{71.2} & 94.3 & 91.6 & 92.9 & \textbf{90.7} & \textbf{71.4} & \textbf{57.9} & \textbf{-3.62} \\
{\small $FT^{init}_{L2}$ }& 79.2 & 76.5 & 73.0 & 95.9 & 95.2 & 94.6 & 97.0 & 95.5 & 92.7 & 83.4 & 80.5 & 75.2 & 73.6 & 67.3 & 94.6 & 91.4 & 91.2 & 88.5 & 63.6 & 44.1 & -4.43\\
{\small $FR$}& 76.7 & 76.6 & 76.4 & 91.7 & 91.5 & 89.7 & 93.8 & 93.5 & 93.5 & 79.5 & 79.4 & 78.7 & 69.2 & 58.2 & 85.8 & 75.2 & 89.3 & 82.8 & 62.3 & 33.5 & -7.62 \\
{\small $DeeSIL$}& 75.5 & 75.1 & 74.3 & 92.7 & 92.5 & 92.2 & 94.0 & 93.7 & 93.2 & 66.9 & 65.8 & 64.2 & 73.0 & 58.1 & 87.2 & 80.0 & 90.5 & 85.1 & 63.9 & 44.0 & -6.92 \\
{\small $REMIND$}& 80.9 & 80.7 & 78.2 & 94.7 & 93.2 & 93.0 & 96.3 & 95.8 & 94.7 & 60.7 & 60.7 & 60.7 & 73.9 & 65.0 & 87.4 & 80.1 & 92.8 & 88.6 & 52.8 & 46.4 & -6.02 \\


\hline 		
$Full$ & \multicolumn{3}{c|}{92.3} &\multicolumn{3}{c|}{99.2} & \multicolumn{3}{c|}{99.1} & \multicolumn{3}{c||}{91.2} & \multicolumn{2}{c|}{92.3} &\multicolumn{2}{c|}{99.2} & \multicolumn{2}{c|}{99.1} & \multicolumn{2}{c||}{91.2} & - \\
\hline
\end{tabular}

}
\end{center}
\vspace{-1em}
	\caption{Top-5 average incremental accuracy (\%) for IL methods with herding using different memory sizes and numbers of IL states. Best results are in bold.
	}
\label{tab:herding_results}
\end{table*}


\begin{table*}
\begin{center}
\resizebox{\textwidth}{!}{
\begin{tabular}{|c||c|c|c|c|c|c|c|c|c|c|c|c||cc|cc|cc|cc||c|}
\hline
States & \multicolumn{12}{c||}{$T=10$} & \multicolumn{8}{c||}{$|\mathcal{K}|=0.5\%$} & \multirow{3}{*}{$G_{IL}$}\\
\cline{1-21}
Dataset & \multicolumn{3}{c|}{ILSVRC} &\multicolumn{3}{c|}{VGGFACE2} & \multicolumn{3}{c|}{LANDMARKS} & \multicolumn{3}{c||}{CIFAR100} & \multicolumn{2}{c|}{ILSVRC} &\multicolumn{2}{c|}{VGGFACE2} & \multicolumn{2}{c|}{LANDMARKS} & \multicolumn{2}{c||}{CIFAR100} & \\
\cline{1-21}
$|\mathcal{K}|$ & $2\%$ & $1\%$ & $0.5\%$ & $2\%$ & $1\%$ & $0.5\%$ & $2\%$ & $1\%$ & $0.5\%$ & $2\%$ & $1\%$ & $0.5\%$ & T=20 & T=50 & T=20 & T=50 & T=20 & T=50 & T=20 & T=50 &   \\
\hline
{\small $iCaRL$ }& 77.9 & 73.0 & 65.3 & 95.3 & 93.8 & 91.1 & 93.9 & 91.4 & 87.4 & 64.5 & 53.4 & 43.9 & 51.3 & 40.9 &  84.3 & 73.2 & 81.9 & 76.4 & 32.6 & 33.4 & -9.51 \\
{\small $LUCIR^{CNN}$ }& 79.8 & 75.9 & 72.2 & \textbf{97.3} & \textbf{97.0} & \textbf{96.6} & 97.1 & 96.4 & 96.0 & 78.6 & 73.9 & 67.5 & 62.4 & 52.9 &  93.5 & 87.8 & 93.2 & \textbf{89.2} & 50.4 & 44.3 & \textbf{-4.36} \\
{\small $FT$} & 77.0 & 70.1 & 60.0& 96.0 & 94.1 & 90.7& 95.8 & 93.2 & 89.1 & 80.0 & 73.7 & 63.3 & 64.5 & 59.2 & 90.8 & 86.5 & 87.8 & 85.5 &  59.9 & 49.4 & -6.40 \\
{\small $BiC$ }& \textbf{85.0} & \textbf{82.4} & \textbf{78.6} & \textbf{97.3} & 96.8 & 96.1 & \textbf{97.8} & \textbf{97.2} & \textbf{96.4} & \textbf{88.2} & \textbf{86.5} & \textbf{82.6} & 72.1 & 59.9 &  92.0 & 82.9 & \textbf{93.8} & 88.1 & 54.2 & 18.1 & -4.42 \\
{\small $ScaIL$ } & 81.0 & 78.2 & 75.1 & 96.4 & 95.6 & 94.5 & 96.9 & 95.3 & 92.7 &  84.6 & 81.1 & 74.9 & \textbf{73.9} & \textbf{68.3} & \textbf{94.5} & \textbf{90.5} & 90.7 & 88.2 & \textbf{67.9} & 47.7 & -4.41 \\
{\small $IL2M$ }& 78.3 & 75.2 & 71.2 & 96.2 & 94.9 & 92.2 & 95.8 & 93.6 & 90.1 & 79.0 & 73.9 & 64.7 & 66.1 & 55.6 & 91.1 & 85.3 & 87.6 & 84.3 & 58.1 & 46.3 & -6.22 \\
{\small $FT^{th}$}& 82.0 & 78.6 & 74.1 & 96.7 & 95.6 & 93.4 & 96.6 & 94.7 & 91.9 & 84.2 & 79.9 & 72.7 & 73.8 & 66.4 & 92.9 & 88.8 & 90.0 & 87.3 & 67.1 & \textbf{52.7} & -4.85 \\
{\small $FR$ }& 74.4 & 74.3 & 74.3 & 83.3 & 83.3 & 83.2 & 93.1 & 93.1 & 92.6 & 78.6 & 78.6 & 78.0 & 66.9 & 54.4 & 76.2 & 49.5 & 84.4 & 71.8 & 58.8 & 28.8 & -12.41 \\
{\small $DeeSIL$}& 74.5 & 74.3 & 74.2 & 92.6 & 92.5 & 92.2& 93.9 & 93.6 & 92.9 & 66.5 & 65.2 & 63.7  & 69.0 & 58.0 &  87.2 & 78.9 &  90.6 & 84.8 & 63.4 & 42.5 & -7.09 \\
\hline 		
$Full$ & \multicolumn{3}{c|}{92.3} &\multicolumn{3}{c|}{99.2} & \multicolumn{3}{c|}{99.1} & \multicolumn{3}{c||}{91.2} & \multicolumn{2}{c|}{92.3} &\multicolumn{2}{c|}{99.2} & \multicolumn{2}{c|}{99.1} & \multicolumn{2}{c||}{91.2} & - \\
\hline
\end{tabular}

}
\end{center}
\vspace{-1em}
	\caption{Top-5 average incremental accuracy (\%) for the main methods with random selection of exemplars for different memory sizes and numbers of IL states. Best results are in bold.
	}
\label{tab:no_herding_results}
\end{table*}

\section{Results and discussion}

\subsection{Incremental learning with memory}
\label{subsec:mem}

Table~\ref{tab:herding_results} presents the performance of all algorithms tested in all experimental configurations when using herding for exemplar selection.  
Both the number of incremental states $T$ and the bounded memory size $|\mathcal{K}|$ have a strong influence on results.
The easiest configurations for all visual tasks are those including a large memory ($|\mathcal{K}|=2\%$) and a small number of states ($T=10$).
Inversely, the most difficult configuration combines a low memory ($|\mathcal{K}|=0.5\%$) and a large number of states ($T=50$).
This finding is intuitive insofar more exemplars for past classes enhance the quality of the replay for them, and a larger number of states makes IL more prone to catastrophic forgetting.
However, the performance drop is more marked for the object recognition tasks (ILSVRC and CIFAR100) compared to face and landmark recognition (VGGFACE2 and LANDMARKS).
For instance, with $T=10$, the ILSVRC's accuracy drop for $BiC$ is of $5.8$ points when moving from $|\mathcal{K}|=2\%$ to $|\mathcal{K}|=0.5\%$ while the corresponding drop for VGGFACE2 is only of $1.6$ points and the one for LANDMARKS is only of $1.3$ points. 
The latter two tasks are simpler, and a smaller amount of exemplars can thus represent past classes.

The increase of $T$, the number of incremental states, also has a detrimental effect on performance.
For fine-tuning based methods, the performance drop is explained by the fact that a larger number of retraining steps causes more information loss, and the effect of catastrophic forgetting is increased.
Also important, for methods like $BiC$, which need a validation set, the size of the latter becomes insufficient when $T$ increases.
This insufficiency is clearly illustrated by $BiC$ results for $|\mathcal{K}|=0.5\%$ and $T=50$.
In this configuration, $BiC$ performance drops more significantly than that of competing methods.
The loss is most striking for CIFAR100, the smallest of all datasets tested, where a performance of only $19.6\%$ is obtained compared to $50.5\%$ for $|\mathcal{K}|=0.5\%$ and $T=20$.
For fixed-representation methods, larger values of $T$ decrease performance because the size of the first non-incremental state becomes smaller.
Consequently, the fixed representation obtained from this state has lower generalization power and is less transferable to later states.

None of the methods is best in all configurations tested.
On aggregate, the best results are obtained with $FT^{th}$, with a $G_{IL} = -3.62$ points loss compared to $Full$, the classical learning upper-bound. 
The other methods with strong performance were all proposed recently: $ScaIL$~\cite{scail2020} ($G_{IL}=-3.7$), $BiC$~\cite{DBLP:conf/cvpr/WuCWYLGF19} ($G_{IL}=-4.03$) and $LUCIR^{CNN}$~\cite{DBLP:conf/cvpr/HouPLWL19} ($G_{IL}=-4.13$). The analysis of individual configurations shows that $BiC$ has good performance in many of them.
This method is best or second-best for the largest memory tested ($|\mathcal{K}|=2\%$) and the smallest number of incremental states ($T=10$).
However, its performance drops faster for the other values of $|\mathcal{K}|$ and $T$.
This is explained by its dependency on a validation set whose size becomes insufficient when $|\mathcal{K}|$ is low, and $T$ is high.

The two $LUCIR$ variants have similar overall performance, with $LUCIR^{CNN}$ being globally better than $LUCIR^{NEM}$. 
This result confirms the original findings reported in~\cite{DBLP:conf/cvpr/HouPLWL19}.
The $iCaRL$ implementation from the same paper ~\cite{DBLP:conf/cvpr/HouPLWL19} has significantly lower performance than the two versions of $LUCIR$.
The positive influence of inter-class separation and cosine normalization introduced in addition to standard knowledge distillation is thus confirmed.

Vanilla $FT$ has lower performance than more recent methods but still much better than $iCaRL$, contrary to the comparison presented in~\cite{DBLP:conf/cvpr/RebuffiKSL17}.
However, the original comparison in that paper was biased since $iCaRL$ used memory while their version of $FT$ was implemented without memory.
All bias reduction methods applied to $FT$ are beneficial, with $FT^{th}$ being the best one followed closely by $ScaIL$. 
$FT^{NEM}$, which exploits the external classifier from~\cite{DBLP:conf/cvpr/RebuffiKSL17} also has interesting performance and outperforms $FT^{BAL}$ and $IL2M$.
The lower performance of the last two methods is an effect of the fact that they are the most sensitive to memory reduction ($|\mathcal{K}|=0.5\%$) and the growth of the number of states ($T=50$). 

$FR$ and $DeeSIL$, the fixed-representation based methods, behave worse than most $FT$-based approaches, with the only exception being that $DeeSIL$ is globally better than $iCaRL$. 
However, it is interesting to note that $FR$ and $DeeSIL$ have a low dependency on memory size, and their performance becomes competitive for $|\mathcal{K}|=0.5\%$. 
In this latter setting, fine-tuning based methods suffer more from catastrophic forgetting since memory becomes insufficient for an efficient replay of past classes.
Globally, $DeeSIL$ has a better behavior compared to $FR$, especially for large scale datasets, and this confirms that the optimization of an external classifier is easier than that of the classification layer of a deep model. 
$REMIND$ performs better than $FR$ and $DeeSIL$ for large datasets and has a better global score ($G_{IL}=-6.02$ VS. $G_{IL}=-7.62$ and $G_{IL}=-6.92$ respectively). However, its performance drops significantly compared to $DeeSIL$ when the number of states is increased from $T=10$ to $T=20$ and $T=50$.

For ILSVRC, $REMIND$ clearly outperforms many FT-based approaches such as $iCaRL$, $LUCIR$, $IL2M$ and $FT$ variants except $FT^{th}$ and $FT^{BAL}$. Streaming based approaches like $REMIND$ have the advantage to run much faster than class incremental based approaches, since they revisit each training example only once. However, the computational cost of $DeeSIL$ is still comparable to that of $REMIND$ since the SVMs training is fast. Equally important, $REMIND$ allows immediate evaluation since it learns the dataset images one by one. It is still usable in class incremental context since we can evaluate the model at the end of all training samples of each incremental state.

\vspace{-0.5em}

\subsection{Role of exemplar selection}
\label{subsec:herding}
As we mentioned, there is an ongoing debate concerning the effectiveness of herding-based versus random exemplar selection in IL~\cite{scail2020,DBLP:conf/eccv/CastroMGSA18,liu2020mnemonics}.
We compare the two selection methods by providing results with random selection for the main algorithms evaluated here in Table~\ref{tab:no_herding_results}.
The obtained results indicate that herding has a positive effect on performance for most of algorithms, albeit with a variable difference with respect to random selection.
The best results in Table~\ref{tab:no_herding_results} are obtained with $LUCIR^{CNN}$, $ScaIL$ and $BiC$ which have very close $G_{IL}$ performance. 
Among fine-tuning based methods, $LUCIR^{CNN}$ and $BiC$ are the methods which are least affected by the switch from herding to random exemplar selection.
Both of these methods implement an end-to-end IL approach.
$iCaRL$ has a more significant performance drop because it makes use of a $NEM$ external classifier.
This is a consequence of the fact that the classifiers are computed directly on the randomly selected exemplars.

Vanilla $FT$ is more affected by the use of random selection than $LUCIR^{CNN}$ and $BiC$. 
The use of distillation for past data partly compensates a poorer class representation with random exemplars. 
Since vanilla $FT$ has a lower performance, algorithms which build on it are also negatively affected. 
Among them, $ScaIL$ is the least affected because it exploits the initial classifiers of past classes. 
$FT^{th}$ performance falls behind that of $LUCIR^{CNN}$, $ScaIL$, and $BiC$ with random exemplar selection because exemplars have a more prominent role for learning past classes' representations.
Thresholding with prior class probabilities is less efficient on poorer past class models.

The use of random selection has a small effect on $FR$ and $DeeSIL$ because the exemplars are only used as negatives when new classifiers are trained. Their presence has a positive effect in that it allows a slightly better separation between new and past classes across IL states. According to the authors of $REMIND$, many herding strategies were deployed based on distance from current example, number of times a sample has been replayed, and the time since it was last replayed, but all of the tested methods performed nearly the same as random selection, with higher computational time.

\subsection{Incremental learning without memory}
\label{subsec:nomem}

\begin{table*}
\begin{center}
\resizebox{0.9\textwidth}{!}{
\begin{tabular}{|c|ccc|ccc|ccc|ccc||c|}
\hline
Dataset & \multicolumn{3}{|c|}{ILSVRC} & \multicolumn{3}{c|}{VGGFACE2} & \multicolumn{3}{c|}{LANDMARKS}& \multicolumn{3}{c|}{CIFAR100} & \multirow{2}{*}{$G_{IL}$}  \\
\cline{1-13}
States & T=10  & T=20  & T=50  & T=10  & T=20 & T=50 & T=10  & T=20 & T=50 & T=10  & T=20 & T=50  & \\
\hline
{\small $LwF$} & 45.3 & 37.6 & 27.1 & 53.3 & 42.6 & 30.8 & 58.8 & 49.2 & 35.2 & 79.5 &  65.3 & 39.0 & -34.72\\
{\small $LwF^{init}$} & 47.1 & 39.9 & 32.2 & 58.1 & 50.8 & 40.5 & 55.7 & 50.2 & 39.8 & 79.4 & 67.9 & 42.8 & -31.97\\
{\small $LwF^{init}_{L2}$ } & 24.5 & 39.7 & 32.0 & 57.1 & 50.7 & 40.5 & 52.1 & 50.5 & 40.0  & 79.5 & 68.1 & 43.3 & -32.60\\
{\small $LwF^{init}_{siw}$} & 54.0 & 45.8 & 35.1 & 70.4 & 59.3 & 45.2 & 61.0 & 53.8 & 42.2 & \textbf{80.0} & \textbf{68.8} & 44.6 & -28.06\\
{\small $LUCIR^{CNN}$} &57.6 & 39.4 & 21.9 & 91.4 & 68.2 & 32.2 & 87.8 & 63.7 & 32.3 & 57.5 & 35.3 & 21.0  & -24.75\\
{\small $FT$} &20.6 & 13.4 & 7.1 & 21.3 & 13.6 & 7.1  & 21.3 & 13.6 & 7.1 & 21.3 & 13.7 & 17.4 & -54.91 \\
{\small $FT^{init}$} &61.0 & 44.9 & 23.8 & 90.9 & 64.4 & 33.1 & 68.8 & 49.4 & 22.2 & 55.1 & 40.8 & 19.9 & -28.99\\
{\small $FT^{init}_{L2}$ } & 51.6 & 43.3 & 34.5 & 76.8 & 66.8 & 55.1 & 61.4 & 52.5 & 39.2  & 47.5 & 39.3 & 22.5 & -26.80\\
{\small $FT^{init}_{L2+mc}$} & 53.6 & 42.7 & 35.6 & 86.9 & 71.4 & 53.6 & 66.2 & 52.6 & 37.9 & 52.6 & 43.1 & 18.2 & -25.02 \\
{\small $FT^{init}_{siw+mc}$ } & 64.4 & 54.3 & 41.4 & 88.6 & 84.1 & 62.6 & 79.5 & 64.5 & 43.2  & 59.7 & 44.3 & 18.4  &  -19.38 \\
{\small $FR$} & 74.0  & 66.9 & 49.2 & 88.7 & 83.0 & 54.4 & \textbf{93.6} & 88.1 & 71.2 &  73.1 & 54.8 & 27.4 & -16.30\\
{\small $DeeSIL$} &\textbf{73.9} & \textbf{67.5} & 53.9 & \textbf{92.3} & \textbf{87.5} & 75.1 & \textbf{93.6} & \textbf{91.1} & \textbf{82.1} & 65.2 & 63.4 & 32.3 & \textbf{-9.22}\\
{\small $REMIND$} & 62.2 & 56.3 & 44.4 & 86.8 & 81.4 & 69.2 & 84.5 & 79.6 & 69.0 & 52.7 & 40.5 & 25.7 & -22.00 \\
{\small $Deep$-$SLDA$} & 70.3 & 64.5 & \textbf{56.0} & 90.2 & 85.4 & \textbf{78.2} & 89.3 & 86.4 & 81.3 & 68.9 & 64.4 & \textbf{54.5} & -15.40\\

\hline
$Full$ & 
\multicolumn{3}{|c|}{92.3}&
\multicolumn{3}{c|}{99.2} & \multicolumn{3}{c|}{99.1} & \multicolumn{3}{c|}{91.2}  & 
- \\
\hline
\end{tabular}
}
\end{center}

	\caption{Top-5 average incremental accuracy (\%) for IL methods without memory for past classes and different numbers of IL states. Best results are in bold.
	}
\label{tab:no_memory_results}
\end{table*}

In many applications, no memory of past classes is available.
For instance, in medical data processing~\cite{DBLP:journals/corr/VenkatesanVPL17}, this is often due to privacy issues.
We study the behavior of the main algorithms which can be deployed in the absence of memory here.
The following algorithms cannot be deployed: (1) all variants which exploit an external $NEM$ classifier since exemplars are not available to build the classifiers for past classes; (2) $BiC$ because it requires a validation set; (3) $ScaIL$ because it requires past exemplars for normalization; (4) $IL2M$ because the mean scores of past classes cannot be computed in the current state.

Without memory, $iCaRL$ becomes $LwF$, the method which inspired more recent works using distillation in IL. 
$LwF^{init}$, $LwF^{init}_{L2}$, and $LwF^{init}_{siw}$ test if the basic hypothesis of $ScaIL$ regarding the reuse of initial classifier weights applies to a method which integrates distillation.
This use of initial weight leads to a 3 points $G_{IL}$ gain compared to classical $LwF$.
Further $L2$ normalization in $LwF^{init}_{L2}$ is not efficient. However, standardization of weights in $LwF^{init}_{siw}$ improves the results of $LwF^{init}$ with 3.9 points.
$LUCIR^{CNN}$ implements a more sophisticated scheme to counter catastrophic forgetting by adding cosine normalization and inter-class separation on top of knowledge distillation.
The two additional components have a significant decisive role since they provide a 10 points gain compared to $LwF$.
Vanilla $FT$ has no component to counter catastrophic forgetting and it has the worst overall performance. 
The use of initial classifiers of past classes in $FT^{init}$ provides a very consequent gain over simple $FT$.
The application of $init$ is much more efficient for $FT$ compared to $LwF$ and even gives nearly the same results than $LwF^{init}_{siw}$, the best variant of $LwF$.
The use of $L2$ normalization in $FT^{init}_{L2}$ improves the results of $FT^{init}$ with 2 $G_{IL}$ points, while adding the mean state calibration $mc$ in $FT^{init}_{L2+mc}$ further gains another 1 $G_{IL}$ points over $FT^{init}_{L2}$.
The best fine tuning based approach without memory is $FT^{init}_{siw+mc}$ from~\cite{siw_20200}. This approach outperforms the other methods with a large margin.

Overall, the best results are obtained with the fixed-representation methods because their dependence on past exemplars is much lower compared to fine-tuning based methods. In order, the best global score is obtained by $DeeSIL$, $Deep$-$SLDA$, $FR$ and $REMIND$.
As we mentioned, $DeeSIL$ is easier to optimize compared to $FR$ and has a comparable accuracy variation with $Deep$-$SLDA$ for most tested configurations. However, $DeeSIL$ provides the best performance when no memory is allowed.
The performance of fixed-representation methods drops when the number of incremental states increases because the initial state includes a lower number of classes.
This is notably the case for CIFAR100, the smallest dataset tested, where the fixed-representations have lower performance compared to all variants of $LwF$ for all tested $T$ values. 
However, $FR$, $REMIND$, $Deep$-$SLDA$ and $DeeSIL$ have consequently better performance for ILSVRC, VGGFACE2, and LANDMARKS where their initial representations are trained with at least 20 classes. 

The analysis of individual datasets shows that $LwF$ variants have a strong performance for CIFAR100, the smallest one among the four tested.
$LwF$ scales worse than $LUCIR^{CNN}$, which is better for the three larger datasets.
The performance inversion is probably explained by the handling of inter-class separation in $LUCIR$.
This indicates that knowledge distillation alone does not scale well because when the number of past classes increases, the confusions between them hamper the performance of the method.
Further analysis of this point is provided in Subsection~\ref{subsec:kd}.

\begin{table}\centering
\resizebox{0.499\textwidth}{!}{
    \begin{tabular}{c|c|c@{\hskip 0.12in}c@{\hskip 0.12in}c@{\hskip 0.12in}c@{\hskip 0.12in}c@{\hskip 0.12in}c@{\hskip 0.12in}c@{\hskip 0.12in}c@{\hskip 0.12in}c@{\hskip 0.12in}c}
        \multicolumn{2}{c|}{Incremental states} & $\mathcal{S}_1$ & $\mathcal{S}_2$  & $\mathcal{S}_3$ & $\mathcal{S}_4$ & $\mathcal{S}_5$ & $\mathcal{S}_6$ & $\mathcal{S}_7$ & $\mathcal{S}_8$ & $\mathcal{S}_9$ \\
        \midrule
        \multicolumn{11}{c}{ILSVRC}\\
        \midrule
         \parbox[t]{2mm}{\multirow{6}{*}{\rotatebox[origin=c]{90}{$LUCIR^{CNN}$}}}   
&$c(p)$  & 62.4 & 46.2 & 36.7 & 29.1 & 23.1 & 18.8 & 15.8 & 14.2 & 13.1 \\
&$e(p, p)$  & 3.8 & 9.6 & 16.2 & 21.3 & 23.6 & 25.2 & 26.0 & 30.2 & 27.9 \\
&$e(p, n)$  & 33.7 & 44.3 & 47.1 & 49.6 & 53.3 & 56.0 & 58.2 & 55.7 & 59.0 \\
&$c(n)$  & 77.9 & 79.2 & 75.2 & 75.2 & 77.9 & 79.4 & 76.9 & 80.5 & 77.7 \\
&$e(n, n)$  & 17.6 & 15.7 & 19.1 & 17.7 & 16.5 & 15.2 & 16.7 & 13.9 & 15.5 \\
&$e(n, p)$  & 4.5 & 5.1 & 5.6 & 7.0 & 5.5 & 5.4 & 6.4 & 5.6 & 6.8 \\

\midrule
\parbox[t]{2mm}{\multirow{6}{*}{\rotatebox[origin=c]{90}{$FT^{init}_{L2}$}}}  
&$c(p)$  & 5.4 & 17.1 & 15.6 & 16.6 & 15.3 & 16.6 & 13.7 & 13.1 & 14.9 \\
&$e(p, p)$  & 0.6 & 12.8 & 10.5 & 28.2 & 20.8 & 47.0 & 27.6 & 28.5 & 55.0 \\
&$e(p, n)$  & 94.0 & 70.1 & 73.9 & 55.2 & 63.8 & 36.4 & 58.8 & 58.4 & 30.1 \\ 
&$c(n)$  & 83.7 & 85.5 & 81.2 & 79.9 & 82.4 & 78.5 & 80.9 & 82.3 & 74.4 \\ 
&$e(n, n)$  & 16.1 & 11.1 & 16.5 & 12.0 & 13.6 & 7.2 & 11.9 & 10.9 & 5.9 \\
&$e(n, p)$  & 0.2 & 3.4 & 2.3 & 8.1 & 4.0 & 14.4 & 7.2 & 6.7 & 19.7 \\



\midrule
\parbox[t]{2mm}{\multirow{6}{*}{\rotatebox[origin=c]{90}{$LwF$}}}
&$c(p)$  & 13.7 & 11.7 & 10.6 & 8.6 & 6.3 & 5.6 & 5.0 & 4.6 & 4.4 \\
&$e(p, p)$  & 6.8 & 17.8 & 25.4 & 25.2 & 27.1 & 28.1 & 32.1 & 33.9 & 33.8 \\
&$e(p, n)$  & 79.6 & 70.5 & 64.0 & 66.1 & 66.6 & 66.4 & 62.9 & 61.5 & 61.7 \\ 
&$c(n)$  & 70.7 & 73.4 & 68.9 & 70.0 & 72.7 & 74.1 & 70.4 & 73.7 & 71.2 \\ 
&$e(n, n)$  & 23.2 & 17.5 & 19.6 & 18.0 & 15.5 & 15.4 & 17.1 & 13.7 & 14.4 \\ 
&$e(n, p)$  & 6.1 & 9.0 & 11.4 & 12.1 & 11.7 & 10.5 & 12.5 & 12.6 & 14.4 \\  

        \hline     
\multicolumn{11}{c}{CIFAR100}\\
\midrule
\parbox[t]{2mm}{\multirow{6}{*}{\rotatebox[origin=c]{90}{$LUCIR^{CNN}$}}}  
&$c(p)$  & 56.7 & 39.4 & 25.7 & 16.8 & 14.4 & 14.5 & 9.4 & 9.8 & 6.7 \\
&$e(p, p)$  & 2.1 & 10.9 & 12.8 & 9.7 & 20.4 & 24.5 & 17.1 & 23.0 & 20.0 \\
&$e(p, n)$  & 41.2 & 49.6 & 61.4 & 73.5 & 65.1 & 61.0 & 73.6 & 67.2 & 73.4 \\ 
&$c(n)$  & 78.9 & 84.4 & 86.9 & 86.3 & 86.5 & 85.3 & 85.0 & 85.2 & 88.2 \\
&$e(n, n)$  & 13.0 & 11.2 & 11.4 & 11.8 & 11.4 & 9.5 & 11.0 & 10.3 & 8.7 \\
&$e(n, p)$  & 8.1 & 4.4 & 1.7 & 1.9 & 2.1 & 5.2 & 4.0 & 4.5 & 3.1 \\

\midrule
\parbox[t]{2mm}{\multirow{6}{*}{\rotatebox[origin=c]{90}{$FT^{init}_{L2}$}}}  
&$c(p)$  & 0.8 & 6.7 & 9.9 & 8.5 & 8.1 & 5.4 & 7.8 & 7.5 & 5.8 \\
&$e(p, p)$  & 0.0 & 4.8 & 9.4 & 10.6 & 25.6 & 12.3 & 23.2 & 38.5 & 19.2 \\ 
&$e(p, n)$  & 99.2 & 88.5 & 80.7 & 80.9 & 66.3 & 82.3 & 69.0 & 54.0 & 75.0 \\ 
&$c(n)$  & 86.7 & 89.2 & 87.8 & 88.2 & 85.2 & 88.1 & 84.0 & 85.3 & 90.7 \\ 
&$e(n, n)$  & 13.2 & 9.5 & 9.2 & 9.1 & 8.9 & 10.2 & 8.9 & 6.1 & 4.6 \\ 
&$e(n, p)$  & 0.1 & 1.3 & 3.0 & 2.7 & 5.9 & 1.7 & 7.1 & 8.6 & 4.7 \\


\midrule
\parbox[t]{2mm}{\multirow{6}{*}{\rotatebox[origin=c]{90}{$LwF$}}}   
&$c(p)$  & 57.3 & 47.5 & 40.3 & 31.1 & 28.7 & 26.6 & 23.8 & 22.0 & 17.7 \\ 
&$e(p, p)$  & 8.0 & 22.6 & 26.2 & 32.1 & 39.8 & 45.5 & 46.5 & 48.5 & 46.0 \\  
&$e(p, n)$  & 34.7 & 29.9 & 33.5 & 36.8 & 31.6 & 27.9 & 29.7 & 29.5 & 36.2 \\ 
&$c(n)$  & 72.7 & 76.4 & 74.4 & 74.9 & 72.6 & 75.4 & 70.6 & 74.6 & 82.3 \\ 
&$e(n, n)$  & 9.9 & 5.1 & 3.9 & 6.9 & 5.0 & 4.0 & 5.6 & 3.2 & 2.4\\  
&$e(n, p)$  & 17.4 & 18.5 & 21.7 & 18.2 & 22.4 & 20.6 & 23.8 & 22.2 & 15.3 \\ 

\midrule
\end{tabular}
}
\caption{Top-1 correct and wrong classification for $LUCIR^{CNN}$, $FT^{init}_{L2}$ and $LwF$ for ILSVRC and CIFAR100 with $T=10$ and $|\mathcal{K}|=0$. $c(p)$ and $c(n)$ are the correct classification for past/new classes. $e(p,p)$ and $e(p,n)$ are erroneous classifications for test samples of past classes mistaken for other past classes and new classes respectively. $e(n,p)$ and $e(n,n)$ are erroneous classifications for test samples of new classes mistaken for past classes and other new classes respectively. Since the number of test images varies across IL states, percentages are calculated separately for test images of past and new classes in each $\mathcal{S}_t$ to get a quick view of the relative importance of each type of errors. $c(p)$, $e(p,p)$, and $e(p,n)$ sum to 100\% on each column, as do $c(n)$, $e(n,n)$, and $e(n,p)$. 
    }
\label{tab:errors}
\end{table}
\vspace{-0.5em}

\subsection{Role of knowledge distillation}
\label{subsec:kd}
In~\cite{DBLP:conf/cvpr/HeZRS16}, authors hypothesize that distillation is useful when the teacher model is trained with a large and balanced dataset.
This is not the case in IL due to the fact that the dataset progressively includes knowledge about more classes and that there is an imbalance between past and new classes. 
In spite of this observation, knowledge distillation is commonly used to tackle catastrophic forgetting~\cite{DBLP:conf/eccv/CastroMGSA18,DBLP:conf/bmvc/He0SC18,DBLP:journals/corr/abs-1807-02802,DBLP:journals/corr/abs-1902-00829,DBLP:journals/corr/abs-1802-07569,DBLP:conf/nips/RebuffiBV17,DBLP:journals/corr/abs-1904-01769}.
Its use in IL with memory was encouraged by the experimental results presented in the influential $iCaRL$ paper~\cite{DBLP:conf/nips/RebuffiBV17}.
There, the original comparison between $FT$ and $iCaRL$ was not fair since the first method is implemented without memory, while the second exploits a memory of the past.
The results reported in Table~\ref{tab:herding_results} for vanilla $FT$ and methods built on top of it challenge the assumption that distillation loss $\mathcal{L}^d$ is necessary in IL with memory.
These experiments show that $FT$ is globally better since the $G_{IL}$ score is over 2 points smaller than that of $iCaRL$.
$iCaRL$ is more effective for ILSVRC and VGGFACE2 datasets only for a small number of incremental states ($T=10$) and the smallest memory ($|\mathcal{K}|=\{1\%,0.5\%\}$).
$FT^{NEM}$ is a version of $iCaRL$ without distillation.
The results from Table~\ref{tab:herding_results} show that the use of the NEM classification layer further improves performance compared to vanilla fine-tuning.

When no memory is allowed for past classes, the experiments reported in Table~\ref{tab:no_memory_results} confirm those presented in~\cite{DBLP:conf/cvpr/RebuffiKSL17}.
There, $LwF$ is clearly better than vanilla $FT$, and the usefulness of distillation is confirmed.
Even without memory, the reuse of the weights of past classes from their initial states in $FT^{init}$ is better than the sole use of knowledge distillation in $LwF$~\cite{DBLP:conf/eccv/LiH16}. 
Only a more sophisticated scheme which combines distillation and an inter-class separation component in $LUCIR^{CNN}$~\cite{DBLP:conf/cvpr/HouPLWL19} outperforms $FT^{init}$, but stays way below $FT^{init}_{siw+mc}$ that does not need distillation. Instead, it makes use of initial weights combined with standardization of all weights and mean state calibration. 

In Table~\ref{tab:errors}, we present the distribution of correct and erroneous predictions across incremental states to have a better understanding of the behavior of distillation in IL.
Results are given for $LwF$~\cite{DBLP:conf/eccv/LiH16}, $LUCIR^{CNN}$~\cite{DBLP:conf/cvpr/HouPLWL19} and $FT^{init}_{L2}$~\cite{scail2020} which implement classical distillation, features-based distillation plus inter-class separation and the reuse of L2-normalized initial classifier weights, respectively.
The authors of~\cite{DBLP:journals/corr/abs-1807-02802} noted that distillation induces a bias among past classes, which leads to confusion between their predictions. 
This finding is confirmed by the large number of past-past class confusions $e(p,p)$ associated to $LwF$ in Table~\ref{tab:errors}.
When advancing in incremental states, the number of past classes increases. 
If an error is made while training the model $\mathcal{M}_1$ using the activations of $\mathcal{M}_0$ as soft targets, it will be passed on to all the subsequent incremental states. 
Consequently, the percentage of $e(p,p)$ errors in Table~\ref{tab:errors} increases in later incremental states. 
However, the percentage of $e(p,p)$ errors is smaller for $LwF$ and $LUCIR^{CNN}$ compared to $FT^{init}_{L2}$ indicating that distillation has a positive effect of past classes.
The addition of the interclass separation in $LUCIR^{CNN}$ removes a part of $e(p,p)$, and the overall increases significantly $c(p)$, the number of correct predictions for past classes.
We note that, since distillation operates on past class scores, the bias in favor of new classes is not handled by $LwF$ and $LUCIR^{CNN}$.
Consequently, $e(p,n)$ is the main type of error for the distillation-based methods.  
This is explained in~\cite{belouadah2019il2m,scail2020, DBLP:conf/cvpr/WuCWYLGF19} by the fact that the model is biased towards new classes, leading to predict past images as belonging to new classes. 
This bias is caused by the fact that new classes are well learned with all their data.

In Table~\ref{tab:no_memory_results}, the comparison of $LwF$, $LUCIR^{CNN}$ and $FT^{init}_{L2}$ for $T=10$ shows that $LwF$ has the lowest and highest performance for ILSVRC and CIFAR100 respectively. 
The detailed view in Table~\ref{tab:errors} gives further insights into the structure of results.
$c(p)$ is higher and $e(p,p)$ is lower for CIFAR100 compared to ILSVRC, indicating that distillation is much more efficient at a smaller scale. 
We conclude that distillation is not always useful in IL. The performance of distilled networks depends on the size of the dataset, the number of incremental states, and the presence or not of the bounded memory of the past. 
It should be used only when the incremental task is known to be characterized by a favorable combination of these parameters.

\subsection{Additional experiment}

\begin{table*}
\begin{center}
    \resizebox{0.95\textwidth}{!}
    {
    \begin{tabular}{|c||c|c|c|c|c|c|c|c|}
        \hline
         IL Method                      & $FT$ & $FT^{th}$ & $iCaRL$ & $BiC$ & $LUCIR^{CNN}$ & $LUCIR^{NCM}$ & \textit{TOPIC-AL} & \textit{TOPIC-AL-MML} \\ \hline
         
         Random  & 23.0 & 32.1 & 48.6 & 49.5 & 55.2 & 61.5 &  \multirow{2}{*}{49.6} & \multirow{2}{*}{49.5}\\
         \cline{1-7}
         Herding   & 27.7 & 38.5 & 51.6 & 52.5 & 55.6 & \textbf{62.6} & &\\ \hline

    \end{tabular}
    }
    \end{center}
    \caption{Top-1 average incremental accuracy for IL methods for IMAGENET100 with 60 initial classes and 8 incremental states containing each 5 classes ($T=9, P_0=60, P_{t=1..8}=5$). The best result is in bold.}
    \label{tab:neural-gas}
\end{table*}

We present a supplementary experiment which compares the performance of a very recent Neural Gas (NG) based approach to that of other methods. 
\textit{TOpology-Preserving knowledge InCrementer}~\cite{fscil_2020} relies on the NG network to preserve the feature space topology using a Hebbian learning~\cite{martinetz1993competitive}. 
Two variants of $TOPIC$ are tested: (1) \textit{TOPIC-AL} - uses an \textit{Anchor Loss} to stabilize the NG network in order to preserve past knowledge and (2) \textit{TOPIC-AL-MML} - uses an \textit{Anchor Loss} and also a \textit{Min-Max Loss} to control the network growth while adapting it to new knowledge.
They are compared to $FT$, $FT^{th}$, $iCaRL$, $BiC$, $LUCIR^{CNN}$, and $LUCIR^{NCM}$.
Note that we use a single dataset because the authors of~\cite{fscil_2020} did not provide their complete code and, while we tried to reproduce their results independently, that was not possible.
Following~\cite{fscil_2020}, we use IMAGENET100~\cite{DBLP:journals/ijcv/RussakovskyDSKS15},  a subset of 100 classes extracted from the ILSVRC dataset, where each class contains 500 training images and 100 test images.
To start with a good data representation, the first model $\mathcal{M}_0$ is trained on $P_0=60$ initial classes. The remaining 40 classes are divided in 8 incremental states containing each $P_t=5$ new classes. The past classes' memory is set to $|\mathcal{K}_{t>0}|=400 + 4 \times (N_t - P_0)$, where 400 images are divided equally between the first state classes, and 4 images per class are used for the classes that do not belong to the first state.
 
Table~\ref{tab:neural-gas} provides top-1 accuracy of classical class IL approaches $FT$, $FT^{th}$, $iCaRL$, $LUCIR$ and $BiC$, and also results of $TOPIC$, the NG-based incremental learner, for IMAGENET100. 
Results indicate that $LUCIR^{NCM}$ is the best approach, followed by $LUCIR^{CNN}$, $BiC$, $iCaRL$, $TOPIC$, $FT^{th}$ and $FT$. 
The two variants of $TOPIC$ provide very similar performances with $TOPIC-AL$ being  marginally better. 
These results are different from those reported in~\cite{fscil_2020}, where $TOPIC$ was found to have better performance.
This difference is probably explained in part by method parametrization choices and in part by the fact that herding was not exploited for $LUCIR$, $BiC$ and $iCaRL$.
We used the original parameters for the methods compared to $TOPIC$ in Table~\ref{tab:neural-gas}.
The use of herding is beneficial for all methods compared to $TOPIC$, with important impact for $FT^{th}$, $iCaRL$ and $BiC$ and lower impact for $LUCIR$ variants.
$TOPIC$ is not affected by the use of herding since this selection is done by the neural gas component.
Globally, the results from Table~\ref{tab:neural-gas} show that, while interesting, the recent adaptation of neural gas approaches to class IL lags well behind the best methods tested in this work.




\section{Conclusion and perspectives}
We presented a comparison of class IL algorithms. 
An analysis based on six desirable properties shows that none of the studied groups is best adapted in all applications. 
We then proposed a formalization of methods that are designed to cope with constant model complexity and with bounded or no memory of past classes.
A selection of recently proposed algorithms is then presented and evaluated thoroughly. 
The evaluation confirms that no algorithm is best in all configurations.

When a memory is allowed, the best global result is obtained when incremental learning is cast as a kind of imbalanced learning.
This type of approach implements a vanilla $FT$ backbone followed by a bias rectification layer. It is especially useful in the most challenging conditions (low memory and many incremental states).
If enough memory is available, and the number of incremental states is low, distillation based approaches become competitive. 
The choice of the method will thus depend on the computation and storage capacities but also on the expected characteristics of the data stream, which needs to be processed.

When no memory is allowed, fixed-representation methods are globally much more competitive than fine-tuning ones while also being simpler and faster to deploy.
They are particularly advantageous for large datasets, where distillation-based methods fail to scale-up. 
This finding is surprising insofar fixed-representation methods exploit a classical transfer learning scheme~\cite{DBLP:journals/corr/abs-1805-08974,DBLP:conf/cvpr/RazavianASC14}.
They do not update models across incremental states and were considered less apt for usage in IL without memory~\cite{DBLP:conf/cvpr/RebuffiKSL17}.
We note that these methods work better than distillation based IL algorithms even when initial representations are learned with a few dozens of classes.

Online learning methods~\cite{DBLP:journals/corr/abs-1910-02509, DBLP:journals/corr/abs-1909-01520} are useful when the stream of data arrives image per image. The model capacity to learn individual classes is increased continuously as new images appear. They are fast to train and well adapted to embedded systems because they have low memory footprint.

For fairness, we evaluated fixed-representation based and fine-tuning based methods with the same initial representation.
If a larger pool of classes is available at the beginning of the process, the performance of fixed-representations will be boosted because the initial representation generalizes better~\cite{deesil}.
However, fixed-representations work well only if the task does not change over time, as it is the case in the evaluated scenarios presented in~\cite{DBLP:conf/eccv/AljundiBERT18}.

The evaluation is done with four different datasets dedicated to distinct visual tasks. 
This setting can be reused and enriched to ensure a robust testing of class incremental algorithms. 
We will release the detailed implementations of all presented methods to facilitate reproducibility.

The comparison presented here shows that recently proposed approaches reduce the performance gap between non-incremental and incremental learning processes.
The analysis of existing algorithms proposed here highlights a series of open problems which could be investigated in the future.
First, handling class IL as an imbalanced learning problem provides very interesting results with~\cite{DBLP:conf/cvpr/WuCWYLGF19} or without~\cite{scail2020} the use of a distillation component. 
Here, we introduced a competitive method where classification bias in favor of new classes is reduced by using prior class probabilities~\cite{DBLP:journals/nn/BudaMM18}.
It would be interesting to investigate more sophisticated bias reduction schemes to improve performance further.
Second, a more in-depth investigation of why distillation fails to work for large scale datasets is needed. 
The empirical findings reported here should be complemented with a more theoretical analysis to improve its usefulness.
Already, the addition of inter-class separation from~\cite{DBLP:conf/cvpr/HouPLWL19} is promising.
More powerful distillation formulations, such as the relational knowledge distillation~\cite{DBLP:conf/cvpr/ParkKLC19}, also hold promise.
Third, the results obtained with herding based selection of exemplars are better compared to a random selection for all methods tested.
Further work in this direction could follow-up on~\cite{liu2020mnemonics} and investigate in more depth which exemplar distribution is optimal for replay.
Finally, the evaluation scenario should be made more realistic by: (1) dropping the strong hypothesis that new data are readily annotated when they are streamed; (2) using a variable number of classes for the incremental states and (3) working with imbalanced datasets, which are more likely to occur in real-life applications than the controlled datasets tested until now.

\textbf{Acknowledgements}. This work was supported by European Union´s Horizon 2020 research and innovation program under grant number 951911 - AI4Media.\\This publication was made possible by the use of the FactoryIA supercomputer, financially supported by the Ile-de-France Regional Council.




\printcredits

\bibliographystyle{cas-model2-names}

\bibliography{cas-refs}


\bio{}
Eden Belouadah holds two Master's, one in Machine Learning and another in Artificial Intelligence and is currently a third-year Ph.D. student working on Large Scale Incremental Learning at IMT Atlantique and CEA LIST, France. She is interested in using Deep Learning to tackle real-world problems, especially those related to computer vision.
\endbio

\bio{}
Adrian Popescu holds a PhD in Computer Science from IMT Atlantique, France (2008) and is currently a researcher with CEA LIST, France. 
His research focuses on the application of machine learning to topics such as: incremental learning, semi-supervised learning, user privacy preservation, and content based image retrieval. 
He is particularly interested in the scalability and practical applicability of the proposed methods. 

\endbio

\bio{}
Ioannis Kanellos is professor at IMT Atlantique, a public engineering school in the French “Grandes Écoles” system. 
He has a background in mathematics and has a core interest in knowledge representation.
He also works on interpretation theories and digital semiotics.
Recently, he contributed to the development of adaptive mediation systems for learning environments and of social robotics for human assisted living.
\endbio

\endbio

\end{document}